\documentclass[a4paper,12pt,twoside=true]{scrbook}

\usepackage[utf8]{inputenc}
\usepackage[T1]{fontenc}
\usepackage[english]{babel}

\usepackage{listing}
\PassOptionsToPackage{export}{adjustbox}
\usepackage{amsfonts}   
\usepackage{amsmath}
\usepackage{amssymb}
\usepackage{stmaryrd}
\usepackage{authblk}
\usepackage{blindtext}
\usepackage[square,sort,comma,numbers]{natbib}
\usepackage[sectionbib, globalcitecopy]{bibunits}
\usepackage{csquotes}
\usepackage{colortbl,xcolor}
\usepackage[hidelinks]{hyperref}
\usepackage{ifthen}
\usepackage{verbatim}
\usepackage{xspace}
\usepackage{tocdata}
\usepackage{dsfont}
\usepackage{chngcntr}
\counterwithout{figure}{chapter}
\counterwithout{table}{chapter}
\counterwithout{equation}{chapter}
\usepackage{cleveref}

\usepackage{svg}
\usepackage{graphicx}
\usepackage{pgfplots}
\pgfdeclarelayer{background}
\pgfdeclarelayer{foreground}
\pgfsetlayers{background,main,foreground} 
\pgfplotsset{compat=1.16}
\usepackage{wrapfig}
\usepackage[mode=buildnew]{standalone} 
\usepackage{tikz}
\usetikzlibrary{automata,arrows,patterns,arrows.meta,positioning,calc,shapes}
\newboolean{externalizer} 
\setboolean{externalizer}{false} 
\newcommand{\ifExtern}[2]{\ifthenelse{\boolean{externalizer}}{#1}{#2}}
\ifExtern{
\usetikzlibrary{external}
\tikzexternalize[]
}{}

\usepackage{multirow, booktabs}
\usepackage{tabu}
\usepackage{tabularx}
\usepackage{makecell}

\usepackage{algorithm} 
\usepackage{algorithmicx}
\usepackage{algpseudocode}

\newcommand{\ie}{i.e.,\ }
\newcommand{\eg}{e.g.,\ }

\newcommand{\setnewpage}{\clearpage\newpage}

\makeatletter
\renewcommand{\sectionauthor}[1]{%
  {\parindent0pt\vspace*{-5pt}%
  \linespread{1.1}\normalfont#1%
  \par\nobreak\vspace*{35pt}}
  \@afterheading%
}
\makeatother

\makeatletter
\renewcommand{\chapterauthor}[1]{%
  {\parindent0pt\vspace*{-15pt}%
  \linespread{1.1}\normalfont#1%
  \par\nobreak\vspace*{35pt}}
  \@afterheading%
}
\makeatother

\newcommand{\VEC}[1]{\mathbf{#1}}          

\newcommand{\PROBD}[1][]{{\mathrm p_{#1}}} 
\newcommand{\PROB}[0]{{\mathrm P}}         
\newcommand{\EXPVAL}{\mathbb{E}}           

\definecolor{tu0}{rgb}{0.7451, 0.1176, 0.2353}

\definecolor{tu1}{rgb}{1.0000, 0.8039, 0.0000}
\definecolor{tu11}{rgb}{1.0000, 0.8627, 0.3020}
\definecolor{tu12}{rgb}{1.0000, 0.9020, 0.4980}
\definecolor{tu13}{rgb}{1.0000, 0.9412, 0.6980}
\definecolor{tu14}{rgb}{1.0000, 0.9608, 0.8000}

\definecolor{tu2}{rgb}{0.9804, 0.4314, 0.0000}
\definecolor{tu21}{rgb}{0.9882, 0.6039, 0.3020}
\definecolor{tu22}{rgb}{0.9882, 0.7137, 0.4980}
\definecolor{tu23}{rgb}{0.9922, 0.8275, 0.6980}
\definecolor{tu24}{rgb}{0.9961, 0.8863, 0.8000}

\definecolor{tu3}{rgb}{0.6902, 0.0000, 0.2745}
\definecolor{tu31}{rgb}{0.7529, 0.2000, 0.4196}
\definecolor{tu32}{rgb}{0.8431, 0.4980, 0.6353}
\definecolor{tu33}{rgb}{0.9216, 0.7490, 0.8196}
\definecolor{tu34}{rgb}{0.9529, 0.8510, 0.8902}

\definecolor{tu4}{rgb}{0.4863, 0.8039, 0.9020}
\definecolor{tu41}{rgb}{0.6431, 0.8627, 0.9333}
\definecolor{tu42}{rgb}{0.7412, 0.9020, 0.9490}
\definecolor{tu43}{rgb}{0.8431, 0.9412, 0.9686}
\definecolor{tu44}{rgb}{0.8980, 0.9608, 0.9804}

\definecolor{tu5}{rgb}{0.0000, 0.5020, 0.7059}
\definecolor{tu51}{rgb}{0.3020, 0.6510, 0.7961}
\definecolor{tu52}{rgb}{0.5490, 0.7765, 0.8667}
\definecolor{tu53}{rgb}{0.7490, 0.8745, 0.9255}
\definecolor{tu54}{rgb}{0.8510, 0.9255, 0.9569}

\definecolor{tu6}{rgb}{0.0000, 0.3255, 0.4549}
\definecolor{tu61}{rgb}{0.2510, 0.4941, 0.5922}
\definecolor{tu62}{rgb}{0.5490, 0.6941, 0.7529}
\definecolor{tu63}{rgb}{0.7490, 0.8314, 0.8627}
\definecolor{tu64}{rgb}{0.8510, 0.8980, 0.9176}

\definecolor{tu7}{rgb}{0.0314, 0.0314, 0.0314}
\definecolor{tu71}{rgb}{0.3725, 0.3725, 0.3725}
\definecolor{tu72}{rgb}{0.5882, 0.5882, 0.5882}
\definecolor{tu73}{rgb}{0.7529, 0.7529, 0.7529}
\definecolor{tu74}{rgb}{0.8667, 0.8667, 0.8667}

\definecolor{tu8}{rgb}{0.7765, 0.9333, 0.0000}
\definecolor{tu81}{rgb}{0.8431, 0.9529, 0.3020}
\definecolor{tu82}{rgb}{0.8863, 0.9647, 0.4980}
\definecolor{tu83}{rgb}{0.9333, 0.9804, 0.6980}
\definecolor{tu84}{rgb}{0.9569, 0.9882, 0.8000}

\definecolor{tu9}{rgb}{0.5373, 0.6431, 0.0000}
\definecolor{tu91}{rgb}{0.6784, 0.7490, 0.3020}
\definecolor{tu92}{rgb}{0.7686, 0.8196, 0.4980}
\definecolor{tu93}{rgb}{0.8588, 0.8941, 0.6980}
\definecolor{tu94}{rgb}{0.9059, 0.9294, 0.8000}

\definecolor{tu10}{rgb}{0.0000, 0.4431, 0.3373}
\definecolor{tu101}{rgb}{0.3020, 0.6118, 0.5373}
\definecolor{tu102}{rgb}{0.5490, 0.7490, 0.7020}
\definecolor{tu103}{rgb}{0.7490, 0.8588, 0.8353}
\definecolor{tu104}{rgb}{0.8549, 0.9176, 0.9059}

\definecolor{tu110}{rgb}{0.8000, 0.0000, 0.6000}
\definecolor{tu111}{rgb}{0.8706, 0.3490, 0.7412}
\definecolor{tu112}{rgb}{0.9216, 0.6000, 0.8392}
\definecolor{tu113}{rgb}{0.9608, 0.8000, 0.9216}
\definecolor{tu114}{rgb}{0.9804, 0.8980, 0.9608}

\definecolor{tu120}{rgb}{0.4627, 0.0000, 0.4627}
\definecolor{tu121}{rgb}{0.5961, 0.2510, 0.5961}
\definecolor{tu122}{rgb}{0.7294, 0.4980, 0.7294}
\definecolor{tu123}{rgb}{0.8392, 0.6980, 0.8392}
\definecolor{tu124}{rgb}{0.9216, 0.8510, 0.9216}

\definecolor{tu130}{rgb}{0.4627, 0.0000, 0.3294}
\definecolor{tu131}{rgb}{0.6118, 0.3020, 0.5333}
\definecolor{tu132}{rgb}{0.7569, 0.5490, 0.6980}
\definecolor{tu133}{rgb}{0.8667, 0.7490, 0.8314}
\definecolor{tu134}{rgb}{0.9216, 0.8510, 0.9020}

\usepackage[headsepline]{scrlayer-scrpage}
\clearpairofpagestyles

\usepackage{caption}
\usepackage[labelformat=simple, labelsep=space]{subcaption}

\newenvironment{abstract}{%
{\large\textbf{Abstract}}\,}
{}

\automark[]{chapter}
\ihead{\normalsize\leftmark}
\ofoot{\thepage}

\makeatletter
\renewcommand{\thesection}{\@arabic\c@section}
\makeatother
\newcommand{\changelocaltocdepth}[1]{%
  \addtocontents{toc}{\protect\setcounter{tocdepth}{#1}}%
  \setcounter{tocdepth}{#1}%
}

\newif\ifchapone
\newif\ifchaptwo
\newif\ifchapthree
\newif\ifchapfour
\newif\ifchapfive
\newif\ifchapsix
\newif\ifchapseven
\newif\ifchapeight
\newif\ifchapnine
\newif\ifchapten
\newif\ifchapeleven
\newif\ifchaptwelve
\newif\ifchapthirteen
\newif\ifchapfourteen
\newif\ifchapfifteen
\newif\ifchapsixteen

\usepackage{layouts}

\chaptwotrue

\begin{document}

\renewcommand\bibname{References}
\newcommand\bibstylenew{halpha_modified}

\ifchapone
\begin{bibunit}[\bibstylenew]
\addchap{Inspect, Understand, Overcome: A Survey of Practical Methods for AI Safety}\label{sec:survey}
\addtocontents{toc}{Sebastian Houben et al.\protect\vspace{0.2em}\par}
\chapterauthor{
Sebastian Houben\footnotemark[1],
Stephanie Abrecht\footnotemark[2],
Maram Akila\footnotemark[1],
Andreas Bär\footnotemark[15],
Felix Brockherde\footnotemark[10],
Patrick Feifel\footnotemark[8],
Tim Fingscheidt\footnotemark[15],
Sujan Sai Gannamaneni\footnotemark[1],
Seyed Eghbal Ghobadi\footnotemark[8],
Ahmed Hammam\footnotemark[8],
Anselm Haselhoff\footnotemark[9],
Felix Hauser\footnotemark[11],
Christian Heinzemann\footnotemark[2],
Marco Hoffmann\footnotemark[16],
Nikhil Kapoor\footnotemark[7],
Falk Kappel\footnotemark[13],
Marvin Klingner\footnotemark[15],
Jan Kronenberger\footnotemark[9], 
Fabian Küppers\footnotemark[9], 
Jonas Löhdefink\footnotemark[15],
Michael Mlynarski\footnotemark[16], 
Michael Mock\footnotemark[1],
Firas Mualla\footnotemark[13],
Svetlana Pavlitskaya\footnotemark[14], 
Maximilian Poretschkin\footnotemark[1], 
Alexander Pohl\footnotemark[16], 
Varun Ravi-Kumar\footnotemark[4], 
Julia Rosenzweig\footnotemark[1],
Matthias Rottmann\footnotemark[5],
Stefan Rüping\footnotemark[1], 
Timo Sämann\footnotemark[4], 
Jan David Schneider\footnotemark[7], 
Elena Schulz\footnotemark[1], 
Gesina Schwalbe\footnotemark[3], 
Joachim Sicking\footnotemark[1], 
Toshika Srivastava\footnotemark[12], 
Serin Varghese\footnotemark[7], 
Michael Weber\footnotemark[14], 
Sebastian Wirkert\footnotemark[6], 
Tim Wirtz\footnotemark[1],
and Matthias Woehrle\footnotemark[2]}
\footnotetext[1]{Fraunhofer Institute for Intelligent Analysis and Information Systems IAIS, sebastian.houben@h-brs.de, \{maram.akila, sujan.sai.gannamaneni, michael.mock, maximilian.poretschkin, julia.rosenzweig, stefan.rueping, elena.schulz, joachim.sicking, tim.wirtz\}@iais.fraunhofer.de}
\footnotetext[2]{Robert Bosch GmbH, \{stephanie.abrecht, christian.heinzemann, matthias.woehrle\}@de.bosch.com}
\footnotetext[3]{Continental AG, gesina.schwalbe@continental.com}
\footnotetext[4]{Valeo S.A., \{varun-ravi.kumar, timo.saemann\}@valeo.com}
\footnotetext[5]{University of Wuppertal, Interdisciplinary Center for Machine Learning and Data Analytics, rottmann@uni-wuppertal.de}
\footnotetext[6]{Bayerische Motorenwerke AG, sebastian.wa.wirkert@bmwgroup.com}
\footnotetext[7]{Volkswagen AG, \{nikhil.kapoor, jan.david.schneider, john.serin.varghese\}@volkswagen.de}
\footnotetext[8]{Opel Automobile GmbH, \{patrick.feifel, seyed.eghbal.ghobadi, ahmedmostafa.hammam\}@opel-vauxhall.com}
\footnotetext[9]{Hochschule Ruhr West, Institute of Computer Science, \{anselm.haselhoff, jan.kronenberger, fabian.kueppers\}@hs-ruhrwest.de}
\footnotetext[10]{umlaut AG, felix.brockherde@umlaut.com}
\footnotetext[11]{Karlsruhe Institute of Technology, felix.hauser@kit.edu}
\footnotetext[12]{Audi AG, toshika.srivastava@audi.de}
\footnotetext[13]{ZF Friedrichshafen AG, \{falk.kappel, firas.mualla\}@zf.com}
\footnotetext[14]{FZI Research Center for Information Technology, \{pavlitskaya, weber\}@fzi.de}
\footnotetext[15]{Technische Universität Braunschweig, Institute for Communcations Technology (IfN), \{andreas.baer, t.fingscheidt, m.klingner, j.loehdefink\}@tu-bs.de}
\footnotetext[16]{QualityMinds GmbH, \{marco.hoffmann, michael.mlynarski, alexander.pohl\}@qualityminds.de}
\setcounter{footnote}{17} 
\input{part1/houben_chap1}\setnewpage
\putbib[references]
\end{bibunit}
\fi


\changelocaltocdepth{0}


\ifchaptwo
\begin{bibunit}[\bibstylenew]
\setcounter{section}{1}
\setcounter{figure}{0}
\setcounter{table}{0}
\setcounter{equation}{0}
\setcounter{algorithm}{0}
\addchap{Does Redundancy in AI Perception Systems Help to Test for Super-Human Automated Driving Performance?}\label{sec:data_requirements}
\addtocontents{toc}{Hanno Gottschalk, Matthias Rottmann, and Maida Saltagic\protect\vspace{0.2em}\par}
\chapterauthor{Hanno Gottschalk\footnotemark[1],
Matthias Rottmann\footnotemark[1],
and Maida Saltagic\footnotemark[2]
}
\footnotetext[1]{University of Wuppertal, School of Mathematics and Science and IZMD, Gaußstr.\ 20, Germany, \{hanno.gottschalk, rottmann\}@uni-wuppertal.de, equal contribution}
\footnotetext[2]{University of Wuppertal, School of Mathematics and Science and IZMD, Gaußstr.\ 20, Germany, maida.saltagic@gmx.de}
\setcounter{footnote}{2} 
\begin{abstract}
While automated driving is often advertised with better-than-human driving performance, this work reviews that it is nearly impossible to provide direct statistical evidence on the system level that this is actually the case. The amount of labeled data needed would exceed dimensions of present day technical and economical capabilities. A commonly used strategy therefore is the use of redundancy along with the proof of sufficient subsystems' performances. As it is known, this strategy is efficient especially for the case of subsystems operating independently, \ie the occurrence of errors is independent in a statistical sense. Here, we give some first considerations and experimental evidence that this strategy is not a free ride as the errors of neural networks fulfilling the same computer vision task, at least for some cases, show correlated occurrences of errors. This remains true, if training data, architecture, and training are kept separate or independence is trained using special loss functions. Using data from different sensors (realized by up to five 2D projections of the 3D MNIST data set) in our experiments is more efficiently reducing correlations, however not to an extent that is realizing the potential of reduction of testing data that can be obtained for redundant and statistically independent subsystems.    
\end{abstract}

\section{Introduction}

The final report of the ethics committee on automated and connected driving \cite{fabio2017ethik} at the German Federal Ministry of Transportation and Digital Infrastructure, starts with the sentences\footnote{translated from German} "Partially and fully automated traffic systems serve first and foremost to improve the safety of all road users. [...]  Protecting people takes precedence over all other utilitarian considerations. The goal is to reduce harm up to complete prevention. The approval of automated systems is only justifiable if, in comparison with human driving performance, they promise at least a reduction of damage in the sense of a positive risk balance".  This pronounced statement sets highest safety goals. In this article, we contemplate the feasibility of a justification based on direct empirical evidence. 

If it comes to automated driving, the elephant in the room is the outrageous amount of data that is needed to empirically support the safety requirement set up by the ethics committee with a direct measurement. This article, however, is not the elephant's first sighting, see, e.g.,\ \cite{kalra2016driving}, where it is shown that hundreds of millions to billions of test kilometers are required for statistically valid evidence on better-than-human driving performance by automated vehicles. 
While in this article the basic statistical facts on the measurement of the probability of rare events are revisited and adapted to a German context, we slightly extend the findings by estimating the data required for testing of AI-based perception functionality using optical sensors along with an estimate of the labeling cost for a sufficient test database.   

What is new in this article is a statistical discussion and preliminary experimental evidence on redundancy as a potential solution to the aforementioned problem. The decomposition of the system into redundant subsystems, each one capable to trigger the detection of other road users without fusion or filtering, largely reduces the amount of data needed to test each subsystem. However, this is only true if failure of the subsystems is statistically independent of the  other subsystems. This leads to the question (a) how to measure independence and (b) whether the actual behavior of neural networks supports the independence assumption. 

A study on the role of independence in ensembles of deep neural networks was presented in \cite{Liu2019independence}, where the goal was rather (1) to improve performance by selecting ensemble members according to different diversity scores and (2) to obtain robustness against adversarial attacks. In \cite{Wu2020EnsembleBench}, a number of different consensus algorithms, i.e., ensemble voting algorithms, are compared according to different evaluation metrics. Also in that work, networks are trained independently and selected afterwards.

In our own studies of independence of the occurrence of error events in the prediction of neural networks, we provide experiments for classification with deep neural networks on the academic datasets EMNIST \cite{emnist}, CIFAR10 \cite{cifar10}, and 3D-MNIST\footnote{\url{https://www.kaggle.com/daavoo/3d-mnist}}. We consider networks with increasing degree of diversity  with respect to training data, architecture, and weight initialization. Even the most diverse networks exhibit Pearson correlation close to $0.60$, which clearly contradicts the hypothesis of independence. Also, re-training committees of up to 5 networks with special loss functions to increase independence between committee members by far does not achieve the $k$-out-of-$n$ performance predicted for independent subsystems \cite{Zacks1992, meeker2014statistical}. 
While it is possible to bring the mean correlation down to zero by special loss functions in the training, this, at least in our preliminary experiments,  at the same time deteriorates the performance of the committee members. As the main take away, redundancy does not necessarily provide a solution to the testing problem.

In this article, we do not aim at presenting final results, but only to provide a contribution to an inevitable debate.   

The remainder of this work is organized as follows: In the next section, we evaluate some numbers from the traffic by motor vehicles in the last pre-pandemic year in Germany, 2019. In Section \ref{sec:Statistical}, we recall some basic facts on the statistics of rare events of an entire system or a system of redundant subsystems. While independent redundant subsystems largely reduce the amount of data required for testing,  we also consider the case of correlated subsystems for which the data requirements scale down less neatly. Also, we discuss the amount of data required to actually prove sufficiently low correlation. In Section \ref{sec:Redundancy}, we test neural networks for independence or correlation for simple classification problems. Not surprisingly, we find that such neural networks actually provide correlated error schemes and the system performance falls far behind the theoretically predicted performance for the error of statistically independent classifiers. This holds true even if we train networks to behave independently or feed the networks with different (toy) sensors. This demonstrates that independence cannot be taken for granted and it might be even hard to achieve through training methods. We give our conclusions and provide a brief outlook on other approaches that have potential to resolve the issue of outrageous amounts of labeled data for a direct assurance case in the final Section \ref{sec:WayOut}.  

\section{How Much Data is Needed for Direct Statistical Evidence of Better-Than-Human Driving?}
\label{sec:BackOfEnvelope}

We focus on the loss of human life as the most important safety requirement. Our frame of reference is set by the traffic in Germany in the year 2019.  For this year, the Federal Ministry of Transport and Digital Infrastructure reports 3,046 fatalities which results in $4.0$ fatalities per billion kilometers driven on German streets in total and $1.1$ fatalities per billion kilometers on motorways, see \cite[p. 165]{BMVI2020verkehr} for these and more detailed data.

If we neglect that some accidents do not involve human drivers, that in deadly accidents oftentimes more than one person is killed and that a large number of those killed did not cause the fatal accident, we obtain a lower bound of at least 250 million kilometers driven per fatality caused by the average human driver. Research on how much this is underestimating the actual distance is recommended but beyond the scope of this work. For an upper bound we multiply this number by an \emph{ad hoc} safety factor of $10$.

Assuming an average velocity in the range of 50 to 100 km/h, this amounts to an average time of about 2.5 to 50 million hours or 285 to 5,700 years of permanent driving until the occurrence of a single fatal accident. If a camera sensor works at a frame rate of 20 to 30 fps,  $(1.8$ to $54)\times 10^{11}$ frames are processed by the AI-system in this time, corresponding to $0.18$ to $5.4$ exabyte ($1$ exabyte $=1\times 10^{18}$ bytes) of data, assuming $1$ megabyte per high resolution image.

Several factors can be identified that would leverage or discount the amount of data required for a direct test of better-than-human safety. We do not claim that our selection of factors is exhaustive and new ideas might change the figures in the future. Nevertheless, here we present some factors that certainly are of importance.

First, due to strong correlation of consecutive frames, the frame rate of $20$ to $30$ fps presumably can be reduced for labeled data. Here, we take the approach that correlation probably is high if the automated car has driven less than one meter, but after $10$ meters driven there is probably not much correlation left that one could infer the safety of the automated car and its environment from the fact that it was safe $10$ m back. At the same time, this approach eliminates the effect of the average traveling speed.   

One could argue further that on many frames no safety-relevant instance of other road users is given and one could potentially avoid labeling such frames. However, from the history of fatal accidents with the involvement of autopilots we learn that such accidents could even be triggered in situations considered to be non-safety-critical from a human perspective, see, e.g., \cite{national2020collision}. As a direct measurement of safety should not be based on assumptions, unless they can be supported by evidence, we do not suggest to introduce a discounting factor as we would have to assume without proof that we could separate hazardous from non-hazardous situations. This of course does not exclude that a refined methodology is developed in the future that is capable to provide this separation and we refer to the extensive literature on corner cases, see, \eg  \cite{bolte:2019,Breitenstein2020,breitenstein2021corner,heidecker2021application}.  

On the other hand, as we will present in Section \ref{sec:Statistical}, a statistically valid estimate on the frequency of rare events requires a leverage factor of at least $3$ to $10$ applied on the average number of frames per incident, see Section \ref{sec:MeasureLowProbabilities} for the details and precise values.  

Further, in the presence of a clear trend of the reduction of fatalities in street traffic \cite{BMVI2020verkehr} (potentially, partially due to AI-based assistance systems already) a mere reproduction of the present day level of safety in driving does not seem to be sufficient. 
Without deeper ethical or scientific justification, we assume that humans expect an at least $10$ to $100$ times lower failure rate of robots than they would concede themselves, while at the same time, we recommend further ethical research and political debate on this critical number. Even with a reduction number of $100$, the approximately $30$ fatalities due to autonomous vehicles would well exceed the risk of being struck by lightning causing $\sim 4$ fatalities per year in Germany\footnote{\url{https://www.vde.com/de/blitzschutz/infos/bitzunfaelle-blitzschaeden\#statistik}}, which often is considered as a generally acceptable risk. 

In addition, several failure modes exist aside AI-based perception that cause fatalities in transportation. We therefore must not reserve the entire cake of acceptable risk to perception-related errors, only. Instead, here we suggest a fraction of $\frac{1}{10}$ to $\frac{1}{2}$ of the entire acceptable risk for perception-related root causes of fatalities. Also at this place, we recommend more serious research, ethical consideration, and public debate. 

Drawing all this together, we obtain a total number of frames that ranges in between $1.50$ trillion frames in the best case scenario to $23{\small,}000$ trillion frames, or $1.5$ to $23{\small,}000$ exabyte (in the year of reference 2019 the entire internet contained $33{\small,}000$ exabyte\footnote{Here, for clarity, we use powers of $10$, e.g., $1000$, instead of powers of $2$, e.g., $1024$.} of data). This computation is summarized from Table \ref{tab:DataRequiremen}.

Note that replacing fatalities with injuries reduces the amount of data by roughly a factor of one hundred (exact number for 2019 $4/509$) \cite{BMVI2020verkehr}. 

A direct measurement of reliability of an AI-perception system requires labeled data and the time to annotate a single frame by a human ranges from a couple of minutes for bounding box labeling up to $90$ minutes for a fully segmented image \cite{schmidt2019crowdproduktion}. Working with the span of $5$ to $90$ minutes per annotated frame and a span of wages from a minimum wage of $9.19$ Euro for Germany in our year of reference 2019 as lower bound to $15$ EUR as upper bound, the cost of labeling a single image ranges between $0.775$ and $22.5$ EUR.

The total cost for labeling of test data to produce direct statistical evidence therefore ranges between $1.16$ trillion in the best case and $51{\small,}800$ trillion Euro in the worst case. This compares to $3.5$ trillion Euro of Germany's gross domestic product in 2019.

We conclude in agreement with \cite{kalra2016driving} that direct and assumption-free statistical evidence of safety of the AI-based perception function of an automated vehicle that complies with the safety requirements derived from the  ethics committee's final report is largely infeasible with the present day technology. 

Certainly, this does not say anything about whether an AI-based system for automated driving actually \emph{would be} driving better-than-human. Many experts, including the authors, believe it \emph{could be}, at least in the long run. But the subjective believe of technology-loving experts ---  in the absence of direct evidence --- is certainly insufficient to give credibility to the \emph{promise} of enhanced safety due to automated driving in the sense of the ethics committee's introductory statement.   

This of course does not exclude that safety arguments which are based on indirect forms of evidence, are reasonable and possible, if they are based on assumptions that can be and are carefully checked. In fact, in the following, we discuss one such potential strategy based on redundancy and report some problems and some progress with this approach applied to small, academic examples.

\begin{table}[ht]
    \caption{Factors and numbers that influence the data requirement for a statistically sound assurance case by direct testing. Safety factors ($\uparrow$) multiply and reduction factors ($\downarrow$) divide the number of frames/cost. }
    \label{tab:DataRequiremen}
\resizebox{\textwidth}{!}{    \centering
    \begin{tabular}{c|c|cc|c|c|cc}\hline
        Description&$\uparrow\downarrow$ &Quantity&Quantity&Unit&Source&Frames &Frames \\
         & & Lower Bound&Upper Bound&  &&Lower Bound&Upper Bound\\\hline\hline
         Meters to fatal  &&$2.50\times 10^{11}$&$2.50\times 10^{12}$&m&\cite{BMVI2020verkehr}&&\\
         accident (2019)&&&&&\& ad hoc &&\\\hline\hline
         Meters driven&$\downarrow$&1&10&m/f&ad hoc&$2.50 \times 10^{10}$&$2.50\times 10^{12}$\\
         per frame&&&&&assumption&&\\\hline\hline
         Factor for&$\uparrow$&2.99&9.21&factor&Section &$7.49\times 10^{10}$&$2.30\times 10^{13}$\\
         stat.~evidence&&$\alpha=5\%$&$\alpha=0.01\%$&&\ref{sec:MeasureLowProbabilities}&&\\\hline\hline
         Add.~safety by&$\uparrow$&10&100&factor&ad hoc &$7.49\times 10^{11}$&$2.30\times 10^{15}$\\
         autom.~driving&&&&&assumption&&\\\hline\hline
         Fraction of perception&$\downarrow$&$\frac{1}{10}$ &$\frac{1}{2}$&factor&ad hoc&$1.50\times 10^{12}$&$2.30\times 10^{16}$\\
         risk from total risk&&&&&assumption&&\\
         \hline\hline
         &&&&&&Cost (EUR)&Cost (EUR)\\
         &&&&&&lower bound&upper bound\\\hline\hline
         Labeling time&$\uparrow$&5&90&min&\cite{schmidt2019crowdproduktion}&&\\
         per frame&&&&& \& ad hoc&&\\\hline\hline
         Hourly wages&$\uparrow$&$9.19$&$15$&EUR/h&minimum wage GER&&\\
         &&&&&2019 \& ad hoc&&\\\hline\hline
         Cost per frame&$\uparrow$&0.775&22.5&EUR/f&2 rows above&&\\\hline\hline
         Total cost &&&&EUR&frames$\times$&$1.16\times 10^{12}$&$5.18\times 10^{17}$\\
         &&&&&cost per frame&&\\\hline
    \end{tabular}}
\end{table}

\section{Measurement of Failure Probabilities}
\label{sec:Statistical}

\subsection{ Statistical Evidence for Low Failure Probability}
\label{sec:MeasureLowProbabilities}

In this subsection we provide the mathematical reasoning for the leverage factor of $2.99$ to  $9.21$ that accounts for statistical evidence. Let us denote by $\PROBD$ the actual probability of a fatal accident for one single kilometer of automated driving. We are looking for statistical evidence that $\PROBD\leq\PROBD_\text{tol}=f_\text{tol}\cdot\PROBD_\text{human}$, where $f_\text{tol}$ is a debit factor for enhanced safety of robots and multiple technical risks. From Table \ref{tab:DataRequiremen} we infer that $f_\text{tol}\in[\frac{1}{1000},\frac{1}{20}]$ taking into account the fraction of perception risk from total risk and the additional safety due to automated driving, cf. Section \ref{sec:BackOfEnvelope}. Here, $\PROBD_\text{human}\approx \frac{1}{250{\small,}000{\small,}000}$ is the (estimated, upper bound) probability of a fatal accident caused by a human driver per driven kilometer. With $\hat p=\frac{N_\text{obs}}{N_\text{test}}$ we denote the estimated probability of a fatal accident per kilometer driven for the autonomous vehicle based on the observed number of fatal accidents $N_\text{obs}$ on $N_\text{test}$ kilometers of test driving. We want to test for the alternative hypothesis $H_1$ that $\PROBD$ is below $\PROBD_\text{tol}$ at a level of confidence $1-\alpha$ with $\alpha\in(0,1)$  a small number, e.g., $\alpha=5\%$, $1\%$, $0.1\%$, $0.01\%$ or even smaller. We thus assume the null hypothesis $H_0$ that $\PROBD\geq \PROBD_\text{tol}$ using that under the null hypothesis $N_\text{obs}\sim B(N_\text{test},\PROBD_\text{tol})$ is Bernoulli distributed with probability $\PROBD_\text{tol}$ and $N_\text{test}$ repetitions. The exact one-sided Bernoulli test rejects the null hypothesis and accepts $H_1$ provided that   
\begin{eqnarray}
\label{eq:BernoulliTest}
\begin{aligned}
    1-\alpha&\leq& \PROB_{N\sim B(N_\text{test},\PROBD_\text{tol})} (N>N_\text{obs})&=1-\PROB_{N\sim B(N_\text{test},\PROBD_\text{tol})} (N\leq N_\text{obs})\\
    &&&=1-\sum_{j=0}^{N_\text{obs}}
    \left(\begin{array}{c}
         N_\text{test}  \\
         j
    \end{array}  \right)
    \left(\PROBD_\text{tol}\right)^j \left(1-\PROBD_\text{tol}\right)^{N_\text{test}-j}.
\end{aligned}
\end{eqnarray}
The reasoning behind \eqref{eq:BernoulliTest} is the following: Assume $H_0$, \ie the true probability of a fatal accident due to the autonomous vehicle would be higher than  $\PROBD_\text{tol}$. Then, with high probability of at least $1- \alpha$ we would have seen more fatal accidents than just $N_\text{obs}$, which we actually observed. This puts us in front of the alternative to either believe that in our test campaign we just observed an extremely rare event of probability $\alpha$, or to discard the hypothesis $H_0$ that the safety requirements are \emph{not} fulfilled.  

Let us suppose for the moment that the outcome of the test campaign is ideal, \ie no fatal accidents are observed at all, \ie $N_\text{obs}=0$. In this ideal case, \eqref{eq:BernoulliTest} is equivalent to
\begin{equation}
    \label{eq:BernoulliBest}
    \alpha\geq \left(1-\PROBD_\text{tol}\right)^{N_\text{test}}~~~\Leftrightarrow~~~ -\frac{\ln(\alpha)}{N_\text{test}}\leq -\ln \left(1-\PROBD_\text{tol}\right)\approx \PROBD_\text{tol},
\end{equation}
where we used the 1st order Taylor series expansion of the natural logarithm at $1$, which is highly precise as $\PROBD_\text{tol}$ is small. Thus, even in the ideal case of zero fatalities observed, $N_\text{test}\geq -\frac{\ln(\alpha)}{\PROBD_\text{tol}}$ is required.  For $\alpha$ ranging between $5\%$ and $0.01\%$, $-\ln(\alpha)$ roughly ranges between $3$ (numerical value $2.9976$) and $10$ (numerical value $9.2103$). This explains the back of the envelope estimates in Section \ref{sec:BackOfEnvelope}.  

Note that the approach of \cite{kalra2016driving} differs as it is based on a rate estimate for the Poisson distribution. Nevertheless, as binominal and Poisson distribution for low probabilities approximate each other very well, this difference is negligible, as the difference is essentially proportional to the event of two or more fatal incidents in one kilometer driven.

\subsection{Test Data for Redundant Systems}
\label{sec:RedndantButCorrelated}

\paragraph{Assuming independence of subsystems:} Let $(\VEC{x},y)$ be a pair of random variables, where $\VEC{x} \in \mathcal{X}$ represents the input data presented to two neural networks $h_1$ and $h_2$, and $y  \in \mathcal{Y}$ denotes the corresponding ground truth label. We assume that $(\VEC{x},y)$ follows a joint distribution $\PROB$ possessing a corresponding density $\PROBD$.
For each neural network, the event of failure is described by
$$
\mathcal{F}_i := \{ h_i(\VEC{x}) \neq  y \} \, , \quad i=1,2,
$$
with $1_{\mathcal{F}_i}$ being their corresponding indicator variables that are equal to one for an event in $\mathcal{F}_i$ and zero else. If and only if we assume independence of the events $\mathcal{F}_i$, we obtain
$$
\EXPVAL[1_{\mathcal{F}_1} \cdot 1_{\mathcal{F}_2} ] = \PROB( \mathcal{F}_1 \cap \mathcal{F}_2 ) = \PROB( \mathcal{F}_1) \cdot \PROB( \mathcal{F}_2 ) = \EXPVAL[1_{\mathcal{F}_1}]  \cdot \EXPVAL[1_{\mathcal{F}_2} ] \, ,
$$
which implies that the covariance fulfills
$$
\mathrm{COV}(1_{\mathcal{F}_1},1_{\mathcal{F}_2}) = \mathbb{E}[1_{\mathcal{F}_1} \cdot 1_{\mathcal{F}_2} ] - \mathbb{E}[1_{\mathcal{F}_1}]  \cdot \mathbb{E}[1_{\mathcal{F}_2} ] = 0 \, .
$$
This is easily extended to $n$ neural networks $h_i(\VEC{x})$, $i\in\mathcal{I}=\{1,\ldots,n\}$ and their corresponding failure sets $\mathcal{F}_i$. Under the hypothesis of independence of the family of events $\mathcal{F}_i$, we obtain
$$
\PROBD_\text{system}=\EXPVAL\left[\prod_{i\in\mathcal{I}}1_{\mathcal{F}_i} \right]=\prod_{i\in \mathcal{I}}\PROB( \mathcal{F}_i )=\prod_{i\in\mathcal{I}}\PROBD_{\text{sub},i},
$$
where $\PROBD_\text{system}$ is the probability of failure of a system of $\# \mathcal{I}=n$ redundant neural networks working in parallel, where failure is defined as all networks being wrong at the same time \cite{meeker2014statistical} and $\PROBD_{\text{sub},i}=\PROB(\mathcal{F}_i)$ is the probability of failure for the $i$-th subsystem $h_i(\VEC{x})$. 

Let us suppose for convenience that the probability for the subsystems $h_i(\VEC{x})$ are all equal, $\PROBD_{\text{sub},i}=\PROBD_\text{sub}$. Then $\PROBD_\text{system}=\PROBD_\text{sub}^n$. In order to give evidence that $\PROBD_\text{system}<\PROBD_\text{tol}$, it is thus enough to provide evidence for $\PROBD_\text{sub}=\PROBD_{\text{sub},i}<\PROBD_\text{tol}^\frac{1}{n}$ for $i\in\mathcal{I}$. subsystem testing to a confidence of $(1-\alpha)$ on the system level requires a higher confidence at the subsystem level, which, by a simple Bonferroni correction \cite{hedderich2016angewandte}, can be  conservatively estimated as $(1-\frac{\alpha}{n})$. Consequently, by \eqref{eq:BernoulliBest} the amount of data for testing the subsystem $h_i(\VEC{x})$ is given by
\begin{equation}
\label{eq:TestSub}
-\frac{\ln\left(\frac{\alpha}{n}\right)}{N_{\text{test},i}}>  \PROBD_\text{tol}^{\frac{1}{n}}~~~~\Leftrightarrow~~~~N_{\text{test},i}> -\frac{\ln\left(\frac{\alpha}{n}\right)}{\PROBD_\text{tol}^{\frac{1}{n}}}.
\end{equation}
As $\PROBD_\text{tol}^{\frac{1}{n}}$ is much larger than $\PROBD_\text{tol}$, the amount of testing data required is radically reduced, even if one employs $n$ separate test sets for all $n$ subsystems. By comparison of \eqref{eq:BernoulliBest} and \eqref{eq:TestSub}, the factor of reduction is roughly
$$
\gamma_n=\frac{N_\text{test}}{n N_{\text{test},i}}=\frac{1}{n\,p_\text{tol}^{1-\frac{1}{n}}\left(1-\frac{\ln(n)}{\ln(\alpha)}\right)}.
$$    
E.g., for $n=2$ in the best case scenario, $\alpha=5\%$, and $f_\text{tol}=\frac{1}{20}$ the reduction factor is $\gamma_2=28{\small,}712$, which reduces the corresponding $1.5\times 10^{12}$ frames to  $52.2$ million frames.
This already is no longer an absolutely infeasible number. For the worst case scenario, $\alpha=0.01\%$ and $f_\text{tol}=\frac{1}{1000}$, the reduction factor is $\gamma_2=232{\small,}502$, resulting in $98.9
$ billion frames, which seems out of reach, but not to the extent of $2.3\times 10^{16}$ frames.

Keeping the other values fixed, $n=3$ even yields a reduction factor of $\gamma_3=974{\small,}672$ in the best case scenario and  $\gamma_3=11{\small,}818{\small,}614$ in the worst case scenario, resulting in $1,54$ million frames in the best and 
$1,95$ billion frames in the worst scenario. These numbers look almost realizable, given the economic interests at stake.

However, the strategy based on redundant subsystems  comes with a catch. It is only applicable, \emph{if} the subsystems are independent. But this is an assumption that is not necessarily true. We therefore investigate, what happens, if the errors of subsystems are not independent.

\paragraph{Assuming no independence of subsystems:} In this case, the covariance of the error indicator variables $1_{\mathcal{F}_i}$ is not equal to zero and can be regarded as a measure of the joint variability for the random variables $1_{\mathcal{F}_i}$. The normalized version of the covariance is the Pearson correlation
$$
\rho(1_{\mathcal{F}_1},1_{\mathcal{F}_2}) = \frac{\mathrm{COV}(1_{\mathcal{F}_1},1_{\mathcal{F}_2})}{ \sigma( 1_{\mathcal{F}_1} ) \cdot \sigma( 1_{\mathcal{F}_2} ) } \in [-1,1],
$$
where $\sigma( 1_{\mathcal{F}_i} )=\sqrt{\PROBD_{\text{sub},i}(1-\PROBD_{\text{sub},i})}$ denotes the standard deviation of $1_{\mathcal{F}_i}$, which is supposed to be greater than zero for $i=1,2$. The correlation measures the linear relationship between the random variables $1_{\mathcal{F}_i}$ and takes values $\pm1$ if the relationship between the random variables is deterministic.  

Let us first consider a system with two redundant subsystems in parallel, $h_i(\VEC{x})$, $i=1,2$, where we however drop the assumption of independence. Then we obtain
\begin{eqnarray}
\label{eq:CorrBernoulli}
\begin{aligned}
\PROBD_\text{system}&=&\EXPVAL[1_{\mathcal{F}_1}\cdot 1_{\mathcal{F}_2}]\\ &=& \mathrm{COV}(1_{\mathcal{F}_1},1_{\mathcal{F}_2})+\EXPVAL[1_{\mathcal{F}_1}]\cdot\EXPVAL[1_{\mathcal{F}_2}]\\
&=&\rho(1_{\mathcal{F}_1},1_{\mathcal{F}_2})\sqrt{\PROBD_{\text{sub},1}(1-\PROBD_{\text{sub},1})}\sqrt{\PROBD_{\text{sub},2}(1-\PROBD_{\text{sub},2})}+\PROBD_{\text{sub},1}\PROBD_{\text{sub},2}.    
\end{aligned}
\end{eqnarray}
Assuming again equal failure probabilities for the subsystems $\PROBD_{\text{sub}}=\PROBD_{\text{sub},1}=\PROBD_{\text{sub},2}$ and using $1-\PROBD_{\text{sub}}\approx1$ as a good approximation as $\PROBD_{\text{sub}}$ for a safe system is very small, we obtain from  \eqref{eq:CorrBernoulli}
\begin{equation}
\label{eq:ApproxPoFCorr}
\PROBD_\text{system}\approx \rho(1_{\mathcal{F}_1},1_{\mathcal{F}_2})\PROBD_{\text{sub}}+\PROBD_{\text{sub}}^2, 
\end{equation}
\ie we can only expect a reduction of the frames needed for testing which is comparable to the case, where statistical independence holds, if $\rho(1_{\mathcal{F}_1},1_{\mathcal{F}_2})$ is of the same small order of magnitude as $\PROBD_{\text{sub}}$. If, \eg we assume an extremely weak correlation of $\rho(1_{\mathcal{F}_1},1_{\mathcal{F}_2})=0.01$, we can essentially neglect the $\PROBD_{\text{sub}}^2$-term as $\PROBD_{\text{sub}}\ll 0.01$ and realize that the reduction factor essentially is  $\frac{1}{\rho(1_{\mathcal{F}_1},1_{\mathcal{F}_2})}=100$, only. Thus, even for such a pretty uncorrelated error scheme, the number of frames required for testing would be lower bounded by $1.5\times 10^{10}$ to  $2.3\times 10^{14}$ frames, even neglecting Bonferroni correction and independent test sets which make up a multiplication factor $B_2=n(1-\frac{\log(n)}{\log(\alpha)})$ yielding $B_2=2.46$ and $B_2=2.15$, respectively. With these effects taken into account, we arrive at $36.9$ billion frames in the best scenario  and   $4.94\times 10^{14}$ frames in the worst, where even the lower number  of frames seems hardly feasible.

A related computation for $n=3$ yields, to leading order using \eqref{eq:ApproxPoFCorr} and approximating the complement of small probabilities with one and neglecting terms of order $\PROBD_{\text{sub}}^2$, we obtain to highest order
\begin{eqnarray}
\label{eq:CorrPof3}
\begin{aligned}
\PROBD_\text{system}&=&\EXPVAL\left[1_{\mathcal{F}_1}\cdot 1_{\mathcal{F}_2}\cdot 1_{\mathcal{F}_3}\right]\\
&\approx&\rho(1_{\mathcal{F}_1\cap 1_{\mathcal{F}_2} },1_{\mathcal{F}_3})\sqrt{\EXPVAL\left[1_{\mathcal{F}_1}\cdot 1_{\mathcal{F}_2}\right]\PROBD_{\text{sub}}}+\EXPVAL\left[1_{\mathcal{F}_1}\cdot 1_{\mathcal{F}_2}\right]\PROBD_{\text{sub}}\\
&\approx& \rho(1_{\mathcal{F}_1\cap 1_{\mathcal{F}_2} },1_{\mathcal{F}_3})\sqrt{\left(\rho(1_{\mathcal{F}_1},1_{\mathcal{F}_2})\PROBD_{\text{sub}}+\PROBD_{\text{sub}}^2\right)\PROBD_{\text{sub}}}\\
&&+ \left(\rho(1_{\mathcal{F}_1},1_{\mathcal{F}_2})\PROBD_{\text{sub}}+\PROBD_{\text{sub}}^2\right)\PROBD_{\text{sub}}\\
&\approx& \rho(1_{\mathcal{F}_1\cap 1_{\mathcal{F}_2} },1_{\mathcal{F}_3})\sqrt{\rho(1_{\mathcal{F}_1},1_{\mathcal{F}_2})}\PROBD_{\text{sub}} \, .
\end{aligned}
\end{eqnarray}
If we thus assume that both correlations $\rho(1_{\mathcal{F}_1\cap 1_{\mathcal{F}_2} },1_{\mathcal{F}_3})$ between the failure of subsystem $h_3(\VEC{x})$ and the composite redundant subsystem from $h_3(\VEC{x})$ and $h_2(\VEC{x})$ are both equal to $0.01$, we obtain a total reduction factor of roughly $\frac{1}{\rho(1_{\mathcal{F}_1\cap 1_{\mathcal{F}_2} },1_{\mathcal{F}_3})\sqrt{\rho(1_{\mathcal{F}_1},1_{\mathcal{F}_2})}}=1{\small,}000$, which still leads to roughly $1.50\times 10^{9}$ - $2.30\times 10^{13}$ frames, even without Bonferroni correction and independent test sets for subsystems. With both taken into account the amount of data ranges between $5,04$ billion frames in the best scenario to  $9.43\times 10^{13}$ frames in the worst, where only the figure obtained in the best case, based on problematic choices, seems remotely feasible. However, in the presence of domain shifts in time and location, it seems questionable if the road of testing weakly correlated subsystems is viable (supposed they \emph{are} weekly correlated). 

We also note that correlation coefficients as low as $\rho=0.01$ are rarely found in nature and in addition it requires empirical testing to provide evidence for a low correlation. The correlations we measure in Section \ref{sec:Redundancy} for the case of simple classification problems miss this low level by at least an order of magnitude, leading to an extra factor of at least 10 in the above considerations.  

\subsection{Data Requirements for Statistical Evidence of Low Correlation}
\label{sec:evidence_low_corr}

In the preceding section we have analyzed that low correlation between sub-systems  efficiently reduces the data required for testing better-than-human safety of autonomous vehicles. However, to achieve this, \eg in the case of two redundant subsystems, the correlation has to be in the order of magnitude of $\PROBD_\text{sub}=\PROBD_\text{tol}^{\frac{1}{2}}$.    
Statistical evidence for such a low level of correlation requires data itself. Let us suppose the ideal situation once more that a correlation coefficient is strictly zero $\rho=0$ and we would like to compute the number of pairs of observations of the random variables $(1_{\mathcal{F}_1},1_{\mathcal{F}_2})$ that is needed to prove that $\rho$ is in the order of magnitude $\PROBD_\text{sub}$, as required for a decent downscaling of the number of test data frames.  In other words, we have to estimate a number of samples needed to provide statistical evidence at a given significance level $\alpha$ that $\rho(1_{\mathcal{F}_1},1_{\mathcal{F}_2})$ is less than $\PROBD_\text{sub}=\PROBD_\text{tol}^{\frac{1}{2}}$.  

As shown by Raymond Fisher and others, see, \eg \cite{loftus1988essence}, the quantity 
$
\hat Z=\frac{1}{2}\log\left(\frac{1+\hat\rho}{1-\hat\rho}\right)
$
is asymptotically normally distributed with expected value $\frac{1}{2}\log(\frac{1+\rho}{1-\rho})=0$ in our case, where we assumed $\rho=0$. The standard deviation is given by $\sqrt{\frac{1}{N_\text{test}-3}}$. Here, $\hat\rho$ stands for the empirical correlation coefficient of the pair of observations \cite{hedderich2016angewandte}. 

A two-sided confidence interval for a given level of confidence $1-\alpha$ for the observed value $\hat z$ of $\hat Z$ thus is given by
\begin{equation}
\label{eq:ConfIntZ}
\hat z_\pm=\hat z\pm z_{1-\frac{\alpha}{2}}\sqrt{\frac{1}{N_\text{test}-3}},
\end{equation}
where $z_{1-\frac{\alpha}{2}}$ is the $1-\frac{\alpha}{2}$ quantile of the standard normal distribution. Transforming back \eqref{eq:ConfIntZ}, we obtain lower and upper bounds
\begin{equation}
\label{eq:ConfIntRho}
\hat \rho_\pm=\frac{\exp(2\hat z_\pm)-1}{\exp(2\hat z_\pm)+1}.
\end{equation}
Under our best case hypothesis $\rho=0$, the boundaries $\hat z_\pm$ of the confidence interval converge to zero, we may apply the $\delta$-rule with the derivative $\left.\frac{d}{dz}\frac{\exp(2z)-1}{\exp(2z)+1}\right|_{z=0}=\left.\frac{4\exp(2z)}{(\exp(2z)+1)^2}\right|_{z=0}=1$. Let us consider the width $W$ of the confidence interval for $\rho$ for the best possible outcome obtained for $\hat z=0$.  By \eqref{eq:ConfIntRho} and the $\delta$-rule, asymptotically for large $N_\text{test}$ it is given by 
$
W=2z_{1-\frac{\alpha}{2}}\sqrt{\frac{1}{N_\text{test}-3}}
$.
Even if this is the case, to infer that $|\rho|\leq\PROBD_\text{sub}$ with confidence $1-\alpha$, one requires, for the case of two subsystems
$
z_{1-\frac{\alpha}{2}}\sqrt{\frac{1}{N_\text{test}-3}}\leq \PROBD_\text{sub}=\PROBD_\text{tol}^{\frac{1}{2}}
$.
If $N_\text{test}$ is large, we can neglect the $-3$ term and obtain for the best case 
\begin{equation}
\label{eq:DareRho}
   N_\text{test} \approx \frac{z_{1-\frac{\alpha}{2}}^2}{p_\text{tol}}. 
\end{equation}
Not unexpectedly, this brings back the bad scaling behavior observed in  \eqref{eq:BernoulliBest} and the related problematic data requirements, which are essentially the same as for the non-redundant, direct approach.  The numbers for $z_{1-\frac{\alpha}{2}}^2 $ for $\alpha=5\%$ \ldots  $\alpha=0.01\%$ range between  $2.706$ and $13.831$ which essentially confirms the range of roughly $3$ \ldots  $10$ for the statistical safety factor obtained from \eqref{eq:BernoulliTest} and \eqref{eq:BernoulliBest}.

\section{Correlation Between Errors of Neural Networks in Computer Vision}
\label{sec:Redundancy}

As of now, deep neural networks (DNNs) for perception tasks are far away from being perfect. Motivated by common practices in reliability engineering, redundancy, i.e., the deployment of multiple system components pursuing the same task in parallel, might be one possible approach towards improving the reliability of a perception system. 

Redundancy can enter into perception systems in many ways. Assume a system setup with multiple sensors, e.g., camera, LiDAR, and Radar. There are multiple options to design a deep-learning-driven perception system processing the different sensors' data. A non-exhaustive list of designs may look as follows:
\begin{enumerate}
    \item Only a single sensor is processed; this is done by a single DNN;
    \item Only a single sensor is processed; this is done by a committee of DNNs;
    \item All sensors are processed by a single-sensor-fusing DNN;
    \item All sensors are processed by a committee of sensor-fusing DNNs;
    \item Each sensor is processed by a separate DNN, the results are fused afterwards;
    \item Each sensor is processed by a committee of DNNs, the results are fused afterwards.
\end{enumerate}
Except for the first design, all other designs incorporate redundancy. Herein, there are two types of redundancy, redundancy via multiple sensors (all pursuing the same task of perceiving the environment) and redundancy via multiple DNNs. 

Certainly, approach one is only eligible for direct testing, see Section \ref{sec:MeasureLowProbabilities} and the same is true for the 'early fusion'  approach 3. All the other approaches could potentially benefit from redundancy, if independence or low corrrelation of the errors can be assumed. Therefore, the degree of independence, which can be understood and quantified as the degree of uncorrelatedness, is a quantity of interest for safety and also for testing, see Section \ref{sec:RedndantButCorrelated}. However, as explained in Section \ref{sec:RedndantButCorrelated}, in order to use redundancy as a part to the solution of the testing problem outlined in Section \ref{sec:BackOfEnvelope}, correlation has to be extremely low.

For the case of simple DNNs processing the same sensor's data, we give evidence that such low correlation in general does not hold.  The evidence we find rather points in the opposite direction that it is hard to obtain correlation that is below 0.5, even if the training datasets and network architecture are kept well separated. On the other hand one could try to train DNNs such that their failure events are uncorrelated.

In this section, we show preliminary numerical results on MNIST (handwritten digits), EMNIST (also containing handwritten letters), and 3D-MNIST for
\begin{itemize}
    \item training DNNs for independence / less correlated failure events;
    \item the role of different sensors (by viewing 3D-MNIST examples from different directions).
\end{itemize}

Although our findings do not directly apply to physically more diverse sensors like Camera, LiDAR and Radar, these preliminary results indicate that independence of DNN-based perception systems cannot be taken simply for granted, even if different sources of data are employed.

\subsection{Estimation of Reliability for Dependent Subsystems}

Most commonly, the so-called active parallel connection of $n$ subsystems is used, wherein the entire system (meaning the ensemble of DNNs) is assumed to be functional iff at least one subsystem (which corresponds to one committee member $h_i$) is functional. 
However, the active parallel connection is not the only decision rule of interest which can be applied to the committee $h_i$, $i=1,\ldots,n$. For instance, considering a pedestrian detection performed by a committee $h_i$ that detects a pedestrian if at least one of the DNNs does so. For increasing $n$ we would expect an increase in false positive detections, therefore facing the typical trade-off of false positives and false negatives. In order to steer this trade-off, we use $k$-out-of-$n$ systems that are functional iff at least $k$ out of $n$ components $h_i$ are functional, i.e., at most $n-k$ components fail. Hence, we are interested in the event
$$
\left\{ \sum_{i=1}^n 1_{\mathcal{F}_i} < n-k \right\} \,
$$
and its probability which is the probability of the $k$-out-of-$n$ system being functional. The reliability of $k$-out-of-$n$ systems can be expressed analytically in terms of the reliability of its components, see also \cite{Zacks1992}. For $k=1$, this boils down to the active parallel connection.

If we assume independence of the failure events $\mathcal{F}_i$ and that all networks fail with equal probability $\PROB(\mathcal{F}_i)=\PROBD_{\text{sub},i}=\PROBD_\text{sub}$, then the probability that at least $k$-out-of-$n$ networks are functional can be calculated via
\begin{equation} \label{eq:chap2_theoretical_reliability}
    P\left( \sum_{i=1}^n 1_{\mathcal{F}_i} < n-k \right ) = \sum_{j=k}^n \left(\begin{array}{c}
         n  \\
         j
    \end{array}  \right)
    (1-\PROBD_\text{sub} )^j \cdot \PROBD_\text{sub}^{n-j} = 1 - F_B(k-1;n,1-\PROBD_\text{sub}),
\end{equation}
where $F_B$ denotes the distribution function of the binomial distribution. This quantity serves as a reference in our experiments.

\subsection{Numerical Experiments}

In this section, we conduct first experiments using the datasets EMNIST, CIFAR10, and 3D-MNIST. The original dataset MNIST \cite{lecun1998mnist} contains 60,000 gray scale images of size 28$\times$28 displaying handwritten digits (10 classes). EMNIST is an extension of MNIST that contains handwritten digits and characters of the same 28$\times$28 resolution. Of that dataset we only considered the characters (26 classes) of which there are 145,600 available. We used 60,000 images for training, 20,000 for validation, and the rest for testing. CIFAR10 contains 60,000 RGB images of size 32$\times$32 categorized into 10 classes (from the categories animals and machines). We used the default split of 50,000 training and 10,000 test examples. A quarter of the training set we reserved for validation. 3D-MNIST contains point clouds living on a $16^3$-lattice. The dataset contains 10,000 training and 2,000 test examples.

We used convolutional DNNs implemented in \texttt{Keras} \cite{chollet2015keras} with simple architectures, if not stated otherwise they contain 2 convolutional layers with 32 and 64 3$\times$3-filters, respectively, each of them followed by a leakyReLU activation and 2$\times$2 max pooling, and finally a single dense layer. For training, we used a batch size of 256, weight decay of $10^{-4}$, and the Adam optimizer \cite{Kingma2015} with default parameters, except for the learning rate. We started with a learning rate of $10^{-2}$, trained until stagnation and repeated that with a learning rate of $10^{-3}$.

\paragraph{Reducing correlations with varying training data, architecture, and weight initializers:}
First, we study to what extent independence can be promoted by varying the training data, architecture and weight initializers in an active parallel system with $n=2$ DNNs. To this end, we split the training data and validation data into two disjoint chunks, consider another network, where we add a third convolutional layer with 128 filters, again followed by leakyReLU and max pooling, and use the Glorot uniform and Glorot normal initializers with default parameters. For the sake of brevity, we introduce a short three-bit notation indicating the boolean truth values corresponding to the questions 
\begin{equation} \label{eq:chap2_booleantriples}
\text{(same data?, same architecture?, same initializer?).}    
\end{equation}
For instance, $101$ stands for two DNNs being trained with the same data, having different architectures, and using the same initializer. We report results in terms of \emph{average accuracy} estimating $\frac{1}{n} \sum_{i=1}^n (1-\PROB(\mathcal{F}_i)) $ and \emph{joint accuracy} estimating  $1-\PROB(\bigcap_{i=1}^n \mathcal{F}_i)$.

\begin{table}[tb]
    \caption{Correlation coefficients and average accuracies for EMNIST and CIFAR10. The configurations read as defined in \eqref{eq:chap2_booleantriples}. All runs have been performed 10 times, all numbers are averaged over these 10 runs and the standard deviation over these 10 runs are given.}
    \label{tab:chap2_independence1}
    \centering
    \scalebox{0.8}{
    \begin{tabular}{c|cc|cc}
        \hline
        & \multicolumn{2}{c|}{EMNIST} & \multicolumn{2}{c}{CIFAR10} \\
        Config. &  $\rho(1_{\mathcal{F}_1},1_{\mathcal{F}_2})$ & avg.~acc. (\%) &  $\rho(1_{\mathcal{F}_1},1_{\mathcal{F}_2})$ & avg.~acc (\%) \\
        \hline\hline
        111 & 0.71±0.01 & 91.17±0.05 & 0.73±0.01 & 72.27±0.45 \\
        110 & 0.71±0.00 & 91.17±0.06 & 0.74±0.01 & 72.19±0.22 \\
        101 & 0.71±0.01 & 91.13±0.04 & 0.74±0.01 & 72.08±0.43 \\ 
        100 & 0.71±0.01 & 91.13±0.05 & 0.74±0.01 & 72.14±0.36 \\ 
        011 & 0.58±0.01 & 89.65±0.14 & 0.66±0.02 & 66.10±0.74 \\ 
        010 & 0.58±0.01 & 89.76±0.08 & 0.65±0.01 & 66.08±0.35 \\ 
        001 & 0.57±0.01 & 89.74±0.07 & 0.66±0.01 & 66.29±0.65 \\ 
        000 & 0.58±0.01 & 89.62±0.13 & 0.65±0.01 & 66.30±0.38 \\
        \hline
    \end{tabular}
    }
    
\end{table}

Table \ref{tab:chap2_independence1} shows in all cases correlation coefficients much greater than zero. Corresponding $\chi^2$ tests with significance level $\alpha=0.05$ in all cases rejected the hypothesis that the DNNs' are independent. Noteworthily, varying the 
initializer or the network architecture barely changes the results while changing the training data seems to have the biggest impact, clearly reducing the correlation coefficient. A reduction in correlation also reduces the average performance of the networks. For the sake of comparability, all networks were only trained with half of the training data, since otherwise the configurations with equal data would have the advantage of working with twice the amount of training data compared to the configuration with different training data.

Next, we study in this setting whether we can achieve at least conditional independence. To this end, we aim at conditioning to the difficulty of the task by conditioning to softmax entropy quantiles. More precisely, we compute the entropy of the softmax distribution of all data points of both networks. We then sum the entropy values for each data point over the two networks and group all examples into 8 equally sized bins according to ascending summed entropy.

\begin{table}[tb]
    \caption{Correlation coefficients $\rho(1_{\mathcal{F}_1},1_{\mathcal{F}_2})$ for different quantiles of softmax entropy computed on EMNIST and CIFAR10. The configurations read as defined in \eqref{eq:chap2_booleantriples}. The experiments have been repeated 10 times, the corresponding standard errors are of the order of 0.01.}
    \label{tab:chap2_independence_entropy1}
    \centering
    \scalebox{0.8}{
    \begin{tabular}{c|cccccccc|cccccccc}
        \hline
        & \multicolumn{8}{c|}{EMNIST} & \multicolumn{8}{c}{CIFAR10} \\
        Entropy bin & 1 & 2 & 3 & 4 & 5 & 6 & 7 & 8 & 1 & 2 & 3 & 4 & 5 & 6 & 7 & 8  \\ \hline\hline
        Config. & \multicolumn{16}{c}{Correlation coefficients} \\
        \hline\hline
        111 & 1.0 & 1.0 & 1.0 & 1.0 & 0.99 & 0.94 & 0.71 & 0.45
        & 1.0 & 0.99 & 0.94 & 0.79 & 0.64 & 0.57 & 0.45 & 0.26 \\
        110 & 1.0 & 1.0 & 1.0 & 1.0 & 0.99 & 0.95 & 0.71 & 0.45
        & 1.0 & 1.0 & 0.96 & 0.79 & 0.67 & 0.55 & 0.44 & 0.31 \\
        101 & 1.0 & 1.0 & 1.0 & 1.0 & 0.99 & 0.94 & 0.71 & 0.43
        & 1.0 & 0.99 & 0.94 & 0.78 & 0.67 & 0.55 & 0.43 & 0.36 \\
        100 & 1.0 & 1.0 & 1.0 & 0.99 & 1.0 & .095 & 0.71 & 0.45
        & 1.0 & 0.99 & 0.95 & 0.79 & 0.64 & 0.51 & 0.46 & 0.36 \\
        011 & 1.0 & 1.0 & 0.99 & 0.98 & 0.90 & 0.71 & 0.40 & 0.28
        & 0.99 & 0.94 & 0.81 & 0.65 & 0.6 & 0.44 & 0.37 & 0.27 \\
        010 & 1.0 & 1.0 & 1.0 & 0.98 & 089 & 0.73 & 0.43 & 0.26
        & 1.0 & 0.95 & 0.82 & 0.64 & 0.5 & 0.44 & 0.40 & 0.25 \\
        001 & 1.0 & 1.0 & 0.98 & 0.96 & 0.90 & 0.73 & 0.40 & 0.25 
        & 0.99 & 0.94 & 0.83 & 063 & 0.54 & 0.43 & 0.37 & 0.27 \\
        000 & 1.0 & 1.0 & 0.99 & 0.98 & 0.90 & 0.72 & 0.42 & 0.27 
        & 0.99 & 0.95 & 0.80 & 0.65 & 0.53 & 0.45 & 0.35 & 0.23 \\
        \hline
    \end{tabular}
    }
\end{table}

Table \ref{tab:chap2_independence_entropy1} shows that the higher the softmax entropy gets, the less the DNNs failures are correlated. This goes down to correlation coefficients of 0.25 for EMNIST and 0.23 for CIFAR10, when considering the softmax entropy bin no.\ 8 with the highest entropy values. Still, $\chi^2$ tests reveal that the correlations are too strong to assume independent failures.

\begin{figure}[!t]
    \centering
    \includegraphics[trim=10 5 10 35,
    clip,
    width=0.7\textwidth]{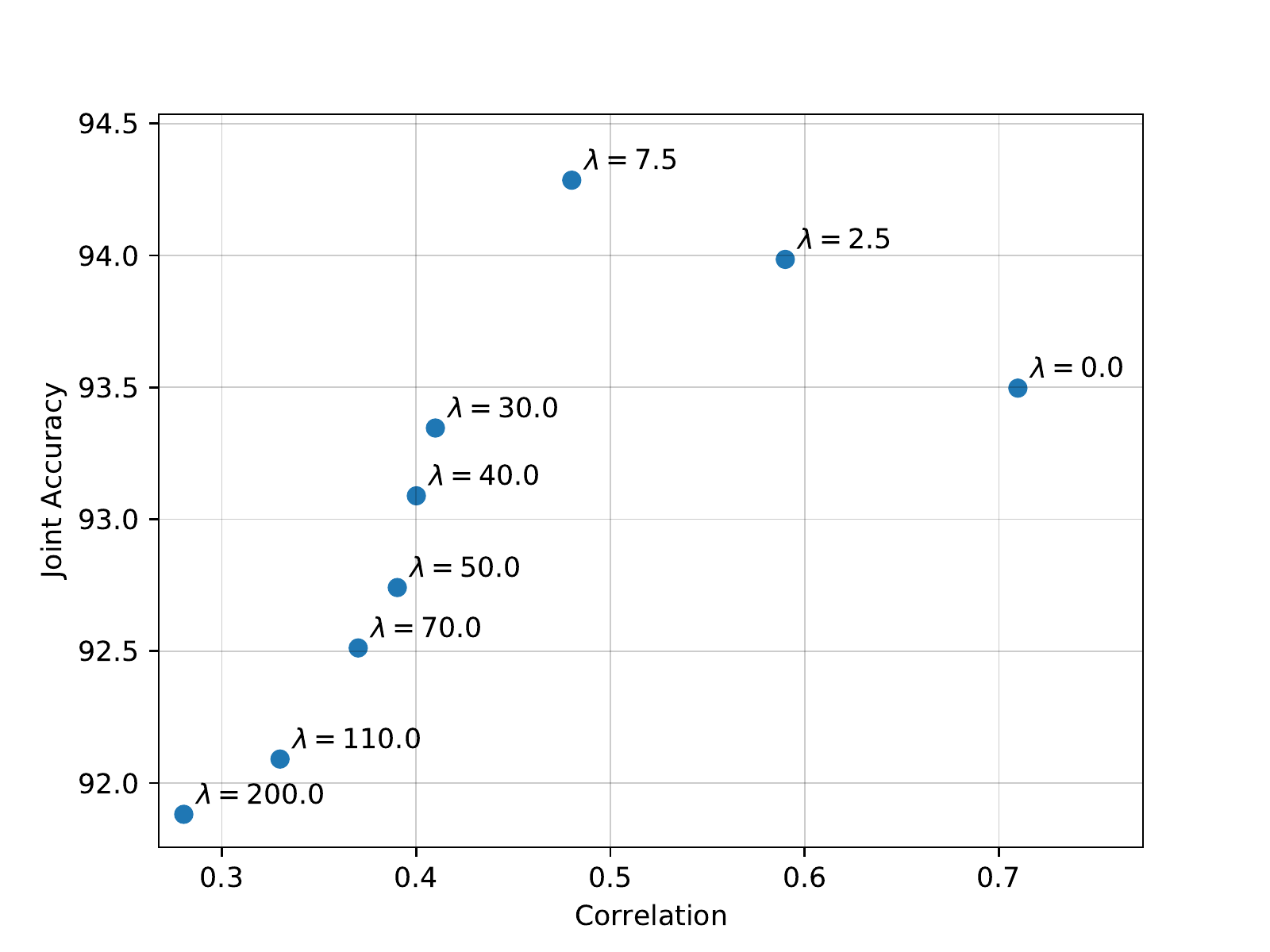}
    \includegraphics[trim=10 5 10 35,
    clip,
    width=0.7\textwidth]{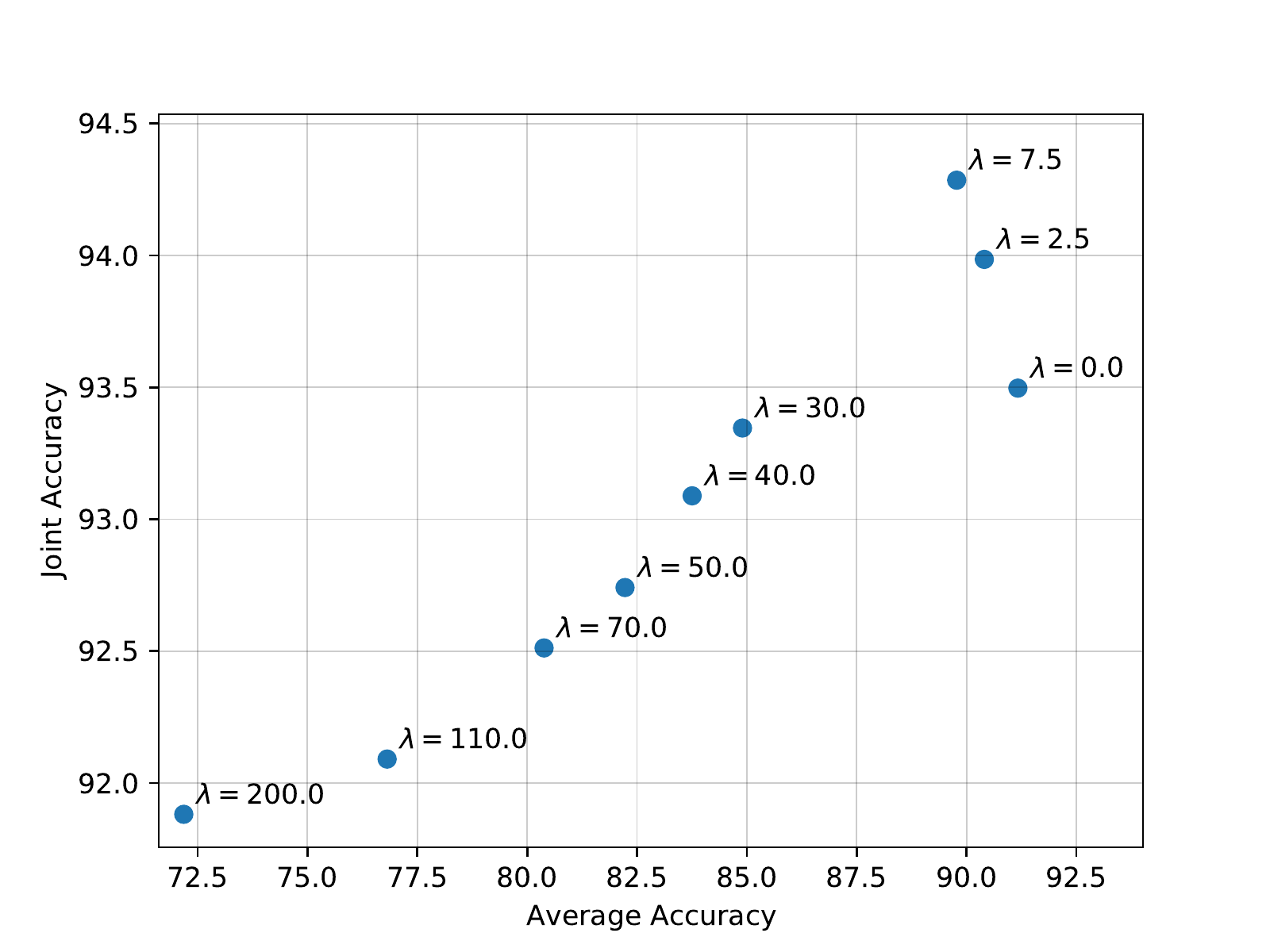}
    \caption{Study of the influence of the loss weight $\lambda$ on averaged accuracy and joint accuracy on the EMNIST dataset.}
    \label{fig:chap2_indepence_training1}
\end{figure}

\paragraph{Training two networks for enhanced independence:}
Since the measures considered so far do not lead to success, we explicitly try to decorrelate the failures of the DNNs. To this end, we incorporate an additional loss function into training, added to the typically used empirical cross-entropy loss. Let $\PROBD( y | \VEC{x}, h_i )$ denote the probability estimated by the DNN $h_i$, that the class $y$ is the correct one given the input $\VEC{x}$. One possible approach is to explicitly enforce a prediction of $h_1$ different to that of $h_2$ if the latter fails and vice versa. This can be achieved by minimizing the following quantity
$$
 - \EXPVAL_{(\VEC{x},y) \sim \PROB} [ 1_{h_i(\VEC{x}) \neq y}  \log( 1 - \PROBD( h_i(\VEC{x}) | \VEC{x}, h_j )) + 1_{h_j(\VEC{x}) \neq y} \log( 1 - \PROBD( h_j(\VEC{x}) | \VEC{x}, h_i ) ]
$$
with its empirical counter part
\begin{align} \label{eq:CCEloss}
J_{i,j}( \{(\VEC{x}_m,y_m)\}_{m=1,\ldots,M} ) =   - \frac{1}{M} \sum_{m=1}^M &  1_{h_i(\VEC{x}_m) \neq y_m}  \log( 1 - \PROBD( h_i(\VEC{x}_m) | \VEC{x}_m, h_j )) \\
& + 1_{h_j(\VEC{x}_m) \neq y_m} \log( 1 - \PROBD( h_j(\VEC{x}_m) | \VEC{x}_m, h_i ) )  \nonumber \, ,
\end{align}
where $\{(\VEC{x}_m,y_m)\}_{m=1,\ldots,M}$ denotes a sample of $M$ data points $(\VEC{x}_m,y_m)$.
In our experiments, we use a penalization coefficient / loss weight $\lambda$ and add $\lambda \cdot J_{1,2}( \{(\VEC{x}_m,y_m)\}_{m=1,\ldots,M} )$ to the empirical cross-entropy loss. In further experiments not presented here, we also used other loss functions that explicitly enforce anti-correlated outputs or even independence of the softmax distributions. However, these loss functions led to uncontrollable behavior of the training pipeline, in particular when trying to tune the loss weight $\lambda$. Therefore, we present results for the loss function in \eqref{eq:CCEloss}.

Figure \ref{fig:chap2_indepence_training1} (top) depicts the correlation as well as the average accuracy for different values of $\lambda$, ranging from 0 to extreme values of 200. For increasing values of $\lambda$, we observe a clear drop in performance from an initial average accuracy of more than $91\%$ for $\lambda=0$ down to roughly $72\%$ for $\lambda=200$. At the same time, the correlation decreased to values below 0.3 which, however, is still not enough to assume independence as being confirmed by our $\chi^2$ tests. While the average accuracy monotonously decreases, it can be observed that the joint accuracy peaks around $\lambda=7.5$, see Figure \ref{fig:chap2_indepence_training1} (bottom). The prediction of the DNN committee is pooled by summing up softmax probabilities over both DNN class-wise and then selecting the class with maximal sum. The joint accuracy is given by the accuracy of the committee with that decision rule. While the joint accuracy for an ordinarily trained committee with $\lambda=0$ is about $93.5\%$, this can be improved by tenderly decorrelating the DNNs to a correlation coefficient around 0.5 which yields an increase of almost 1 percent point. At the same time there is a mild decrease in average accuracy.

\begin{table}[tb]
    \caption{Correlation coefficients under independence training for different quantiles of softmax entropy computed on EMNIST and CIFAR10.}
    \label{tab:chap2_independence_entropy2}
    \centering
    \scalebox{0.8}{
    \begin{tabular}{c|cccccccc|cccccccc}
        \hline
        & \multicolumn{8}{c|}{EMNIST} & \multicolumn{8}{c}{CIFAR10} \\
        Entropy bin & 1 & 2 & 3 & 4 & 5 & 6 & 7 & 8 & 1 & 2 & 3 & 4 & 5 & 6 & 7 & 8  \\ \hline\hline 
        $\lambda$ & \multicolumn{16}{c}{Correlation coefficients} \\
        \hline \hline
        7.5 & 1.0 & 1.0 & 0.99 & 0.96 & 0.90 & 0.59 & 0.35 & 0.07
        & 0.89 & 0.76 & 0.63 & 0.53 & 0.47 & 0.39 & 0.32 & 0.21 \\
        0.0 & 1.0 & 1.0 & 1.0 & 1.0 & 0.99 & 0.95 & 0.72 & 0.44
        & 1.0 & 0.99 & 0.94 & 0.80 & 0.64 & 0.57 & 0.42 & 0.33 \\
        \hline
    \end{tabular}
    }
\end{table}

Concluding this section, we again study conditional independence for 8 softmax entropy bins (chosen according to equally distributed softmax entropy quantiles), this time comparing independence training $\lambda=7.5$ with ordinary training $\lambda=0$, see Table~\ref{tab:chap2_independence_entropy2}. We observe a considerable decrease in correlation in bin no.\ 8 (representing the highest softmax entropy) for the EMNIST data down to 0.07. In comparison, the decrease for CIFAR10 is rather mild. However, also this small correlation coefficient of 0.07 is not sufficient for assuming conditional independence.

\begin{table}[t!]
    \caption{Correlation coefficients for an ensemble of 5 networks trained on EMNIST for independence and a baseline ensemble where each network was trained individually. All results are averaged over 30 runs.}
    \label{tab:chap2_fivenetworks}
    \centering
    \scalebox{0.8}{
    \begin{tabular}{c|c|ccccc|c|c}
        \hline
        Loss weight $\lambda$ & & 0 & $10^{-1}$ & $10^{0}$ & $10^{1}$ & $10^{2}$ & Baseline model & Theoretical \\
        \hline\hline
        Mean correlation & & 0.70 & 0.69 & 0.65 & 0.53 & 0.43 & 0.67 & 0.00\\
        \hline \hline
        \multirow{5}{1.8cm}{Single network accuracy} & $k=1$ & 0.92 & 0.92 & 0.91 & 0.88 & 0.73 & 0.91 \\
        & $k=2$ & 0.92 & 0.92 & 0.91 & 0.91 & 0.91 & 0.91 \\
        & $k=3$ & 0.92 & 0.92 & 0.91 & 0.88 & 0.60 & 0.91 \\
        & $k=4$ & 0.92 & 0.92 & 0.91 & 0.87 & 0.51 & 0.91 \\
        & $k=5$ & 0.92 & 0.92 & 0.91 & 0.85 & 0.39 & 0.91 \\
        \hline\hline
        \multirow{5}{1.8cm}{Ensemble accuracy / Mean $k$-out-of-5 accuracy} & $k=1$ & 0.92 / 0.96 & 0.92 / 0.96  & 0.91 / 0.96 & 0.88 / 0.97 & 0.73 / 0.93 & 0.91 / 0.96 & 1.00 \\
        & $k=2$ & 0.92 / 0.94 & 0.92 / 0.94 & 0.92 / 0.94 & 0.92 / 0.93 & 0.91 / 0.78 & 0.92 / 0.94 & 0.99 \\
        & $k=3$ & 0.92 / 0.92 & 0.92 / 0.92 & 0.92 / 0.92 & 0.92 / 0.90 & 0.90 / 0.64 & 0.92  / 0.92 & 0.96 \\
        & $k=4$ & 0.93 / 0.90 & 0.93 / 0.90 & 0.92 / 0.89 & 0.92 / 0.85 & 0.90 / 0.49 & 0.92 / 0.90 & 0.72 \\
        & $k=5$ & 0.93 / 0.86 & 0.93 / 0.86 & 0.92 / 0.84 & 0.92 / 0.76 & 0.90 / 0.30 & 0.92 / 0.85 & 0.32 \\
        \hline
    \end{tabular}
    }           
\end{table}

\begin{figure}[ht]
    \centering
    \includegraphics[trim=20 0 40 0, clip,width=0.49\textwidth]{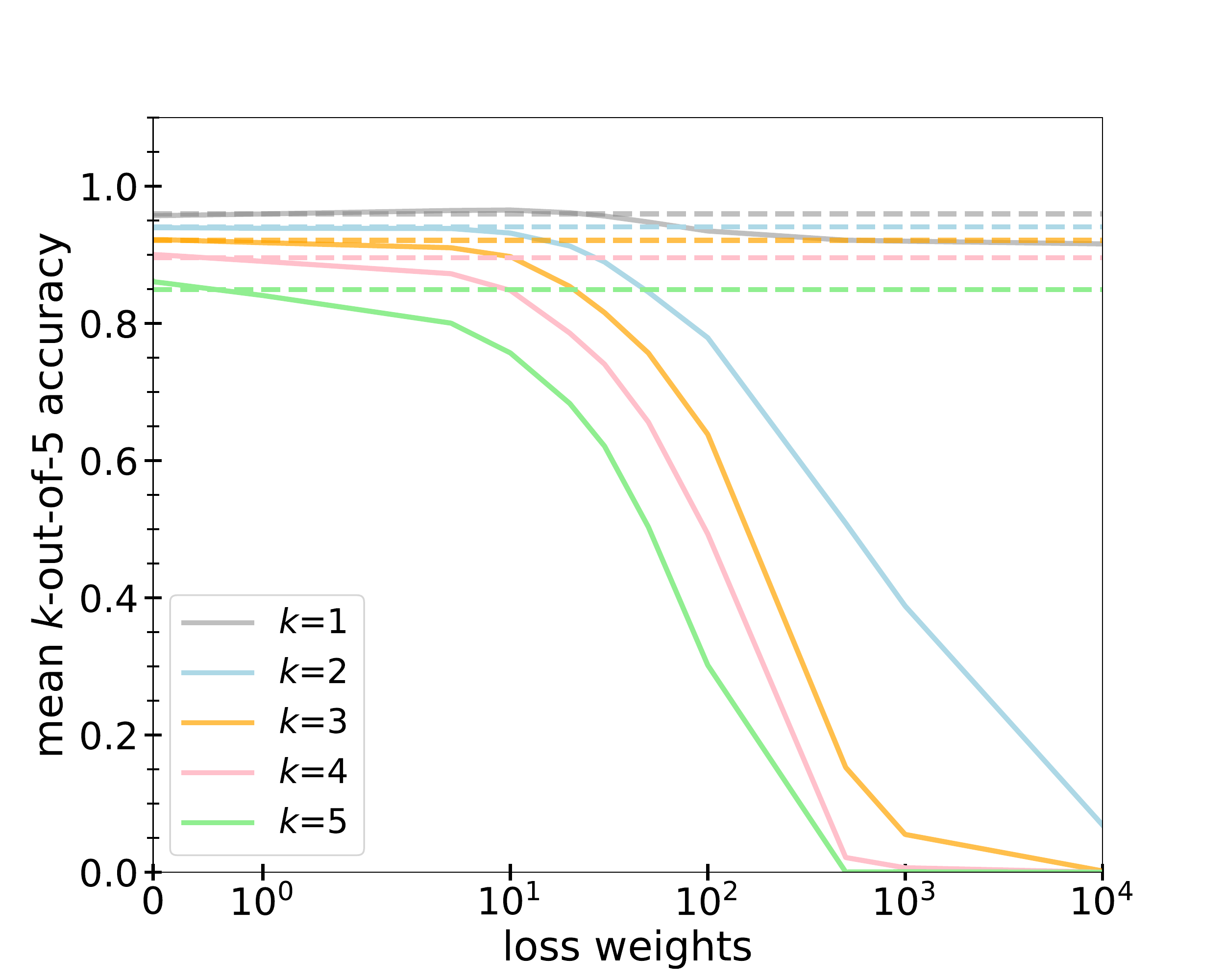}
    \includegraphics[trim=20 0 40 0, clip, width=0.49\textwidth]{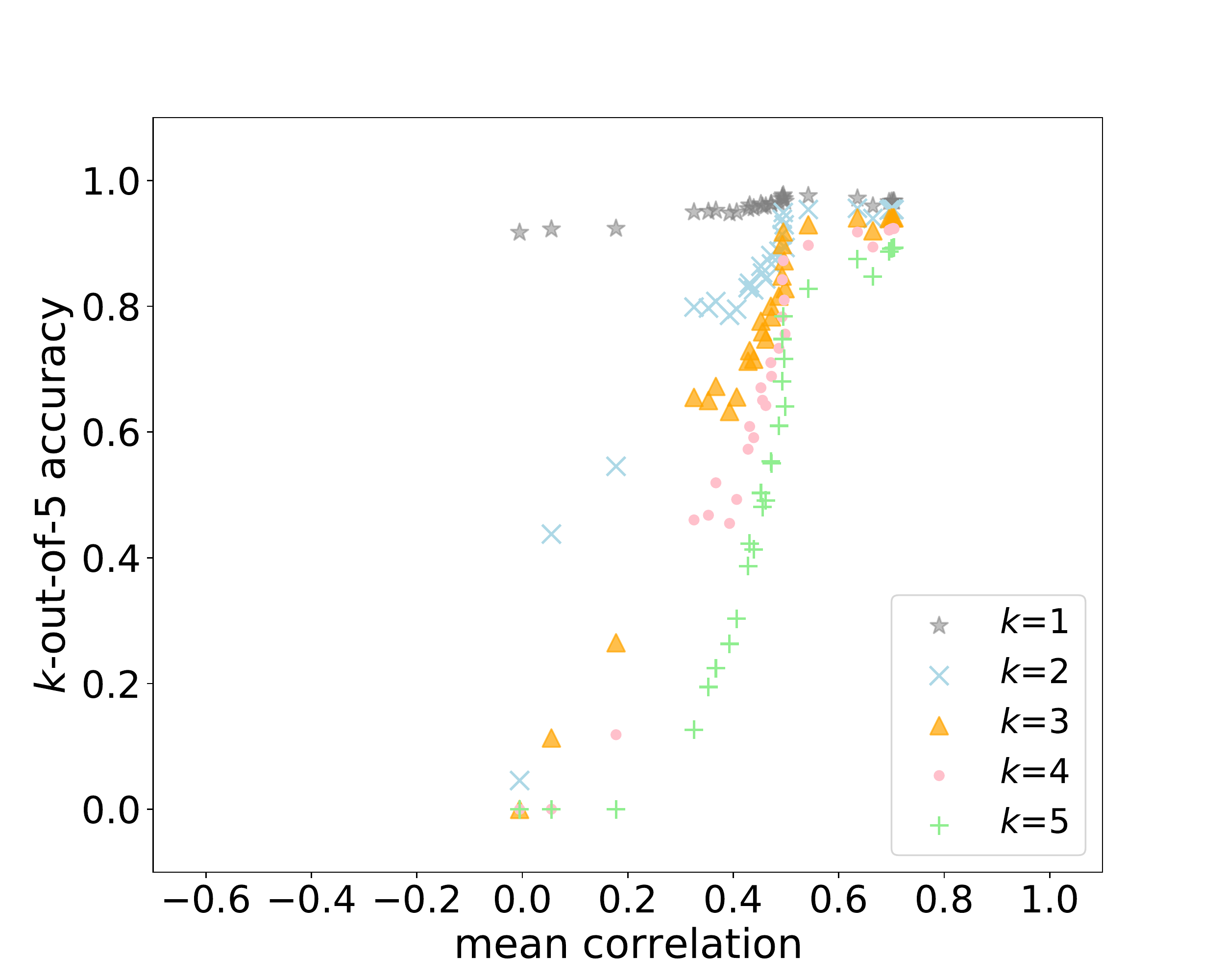}
    \caption{Experiments with the EMNIST dataset. Left: mean $k$-out-of-$5$ accuracy (averaged over 30 repetitions) for an ensemble of five networks as a function of the loss weight $\lambda$. The solid lines depict the accuracies of the ensembles trained for independence with loss weight $\lambda$, the dashed lines depict baseline ensemble accuracies trained without incorporating the loss from ~\eqref{eq:chap2_n_networks_inpendence_loss}. Right: $k$-out-of-$5$ accuracy for a single run as a function of the mean correlation (of each network with each other network).}
    \label{fig:chap2_fivenetworks1}
\end{figure}

\begin{figure}[ht]
    \centering
    \includegraphics[trim=5 0 5 0, clip,width=0.49\textwidth]{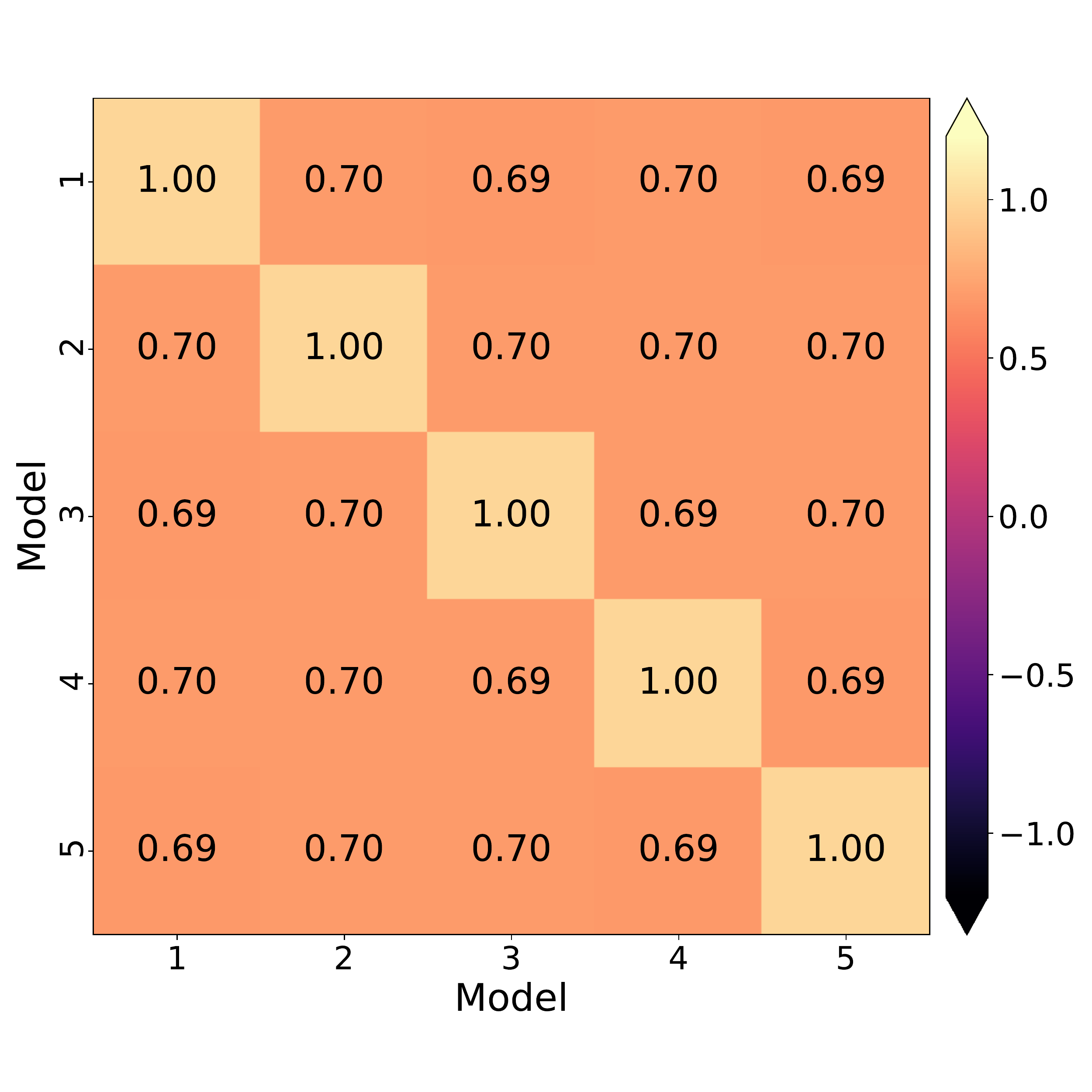}
    \includegraphics[trim=5 0 5 0, clip, width=0.49\textwidth]{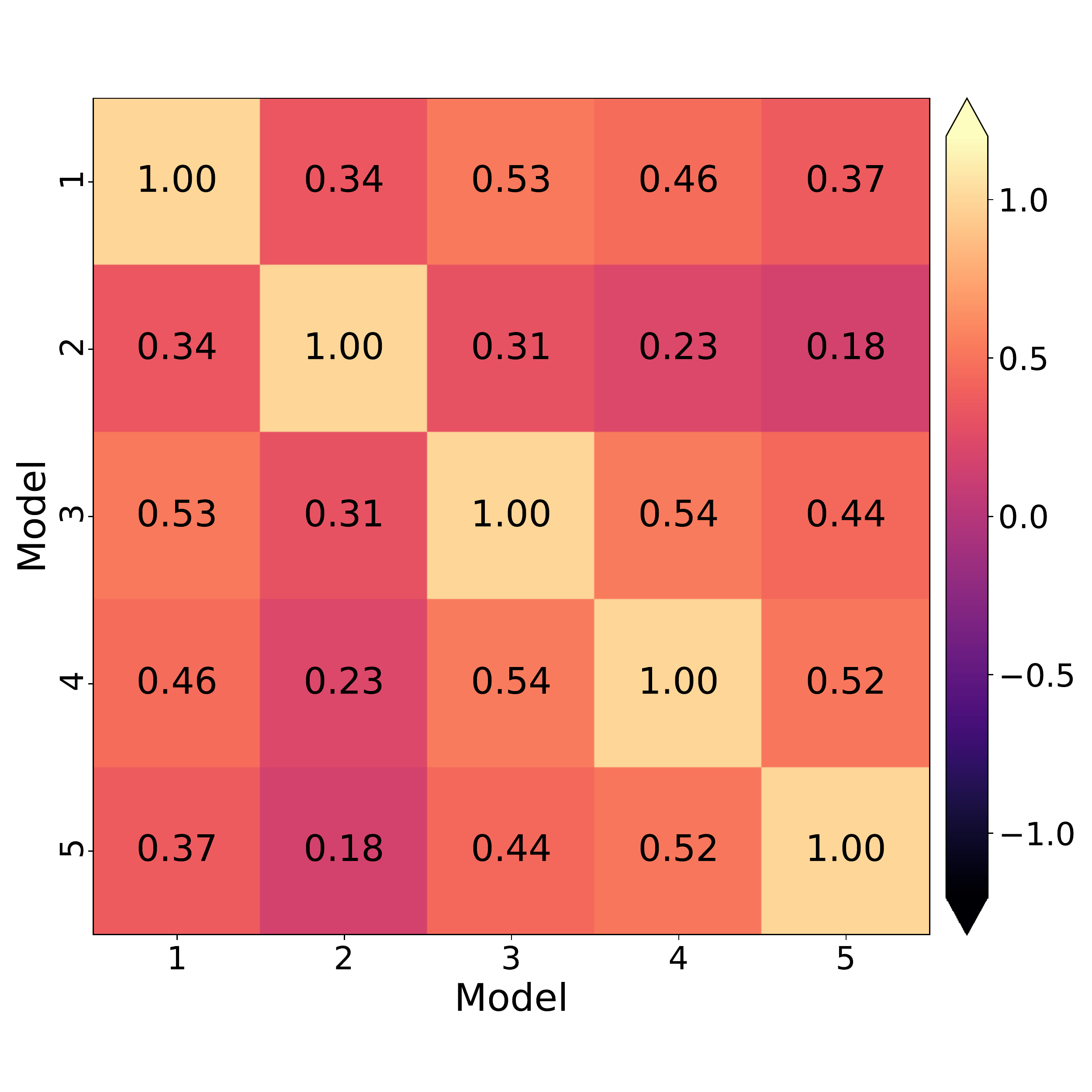}
    \caption{Correlation coefficients of the networks' errors for all combinations of 2 out of 5 models. Left: loss weight $\lambda=10^{-1}$. Right: loss weight $\lambda=10^{2}$.}
    \label{fig:chap2_corr_coeff}
\end{figure}

\paragraph{Training several networks for independence:}
We now present results when training an ensemble of $n=5$ networks. To this end, we sum up the cross-entropy losses of the $n=5$ models and add the penalization term
\begin{equation} \label{eq:chap2_n_networks_inpendence_loss}
\lambda \cdot \frac{2}{n-1} \sum_{i=2}^n \sum_{j=1}^{i-1} J_{i,j}( \{ (\VEC{x}_m,y_m) \}_{m=1\ldots,M} ) 
\end{equation}
which is a straight forward combinatorial generalization of the previously used penalty term. Therein, $\lambda$ again denotes the loss weight which varies during our experiments.

Besides $k$-out-of-$5$ accuracies, we consider also accuracies from single networks and ensemble accuracies. The corresponding ensemble prediction is obtained by summing up the softmax probabilities (via the class-wise sum over the first $k$ ensemble members) and then taking the argmax.

Figure \ref{fig:chap2_fivenetworks1} depicts results of our training for independence with 5 models in terms of $k$-out-of-$5$ accuracy. When stating mean accuracy, this refers to the average over 30 runs. For a loss weight of $\lambda=10^1$, we observe in the left panel that the $1$-out-of-$5$ accuracy of our independence training is slightly above the accuracy of the baseline ensemble which was trained regularly, \ie each network was trained independently. Note that this is different to $\lambda=0$, where all networks are still trained jointly with a common loss function which is the sum of the cross-entropy losses. The right-hand panel shows that the $1$-out-of-$5$ accuracy peaks at a mean correlation (the average correlation over the errors of all networks $i < j$) of $0.4$. Decorrelating the networks' errors further towards zero-mean correlation is possible, however the $1$-out-of-$5$ accuracy decreases. The $k$-out-of-$5$ accuracy for $k>1$ suffers even more from decorrelating the networks' errors. Note that, in practice, $1$-out-of-$5$, e.g., for a pedestrian detection, might additionally suffer from overproduction of false positives and could be impractical. That hypothesis is indeed supported by Table~\ref{tab:chap2_fivenetworks}, which shows that for larger loss weights $\lambda \geq 1$ the individual network accuracies become heterogeneous. In particular, network 5 suffers from extremely low accuracy at $\lambda=10^2$, which is, however, still far away from zero-mean correlation. For the sake of completeness, we give two examples of correlations between the individual models since we only reported mean correlations so far, see Figure~\ref{fig:chap2_corr_coeff}. Comparing the right hand panel with Table~\ref{tab:chap2_fivenetworks}, we see that the well-performing network no.~$2$ exhibits comparatively small correlation coefficients with the other networks' errors. Surprisingly, the worse performing models' errors show higher correlation coefficients which reveals that in that case they often err jointly on the same input examples.
Additionally, we provide theoretical $k$-out-of-$5$ accuracies according to \eqref{eq:chap2_theoretical_reliability} where we choose $\PROBD_\mathrm{sub}$ equal to $1$ minus the average ensemble accuracy (which is $92.45\%$). In particular, the $k$-out-of-$5$ accuracies of the ensemble for $k=1$ and $2$ are clearly below the theoretical ones.

We conclude that independence training may help slightly to improve the performance, however, the benefit seems to be limited when all networks are trained with the same input. Besides these tests, we considered an additional loss function that explicitly penalizes the correlation of the networks' errors. That approach, being actually more directed towards our goal, even achieved negative correlation coefficients of the networks' errors. It was also able to slightly improve the ensembles' performance in terms of $k$-out-of-$n$ accuracy over the baseline, however, this improvement was even less pronounced than that one reported in this section. Thus, we do not report those results here.

\begin{figure}[t!]
    \centering
    \includegraphics[trim=80 40 80 40, clip,width=0.27\textwidth]{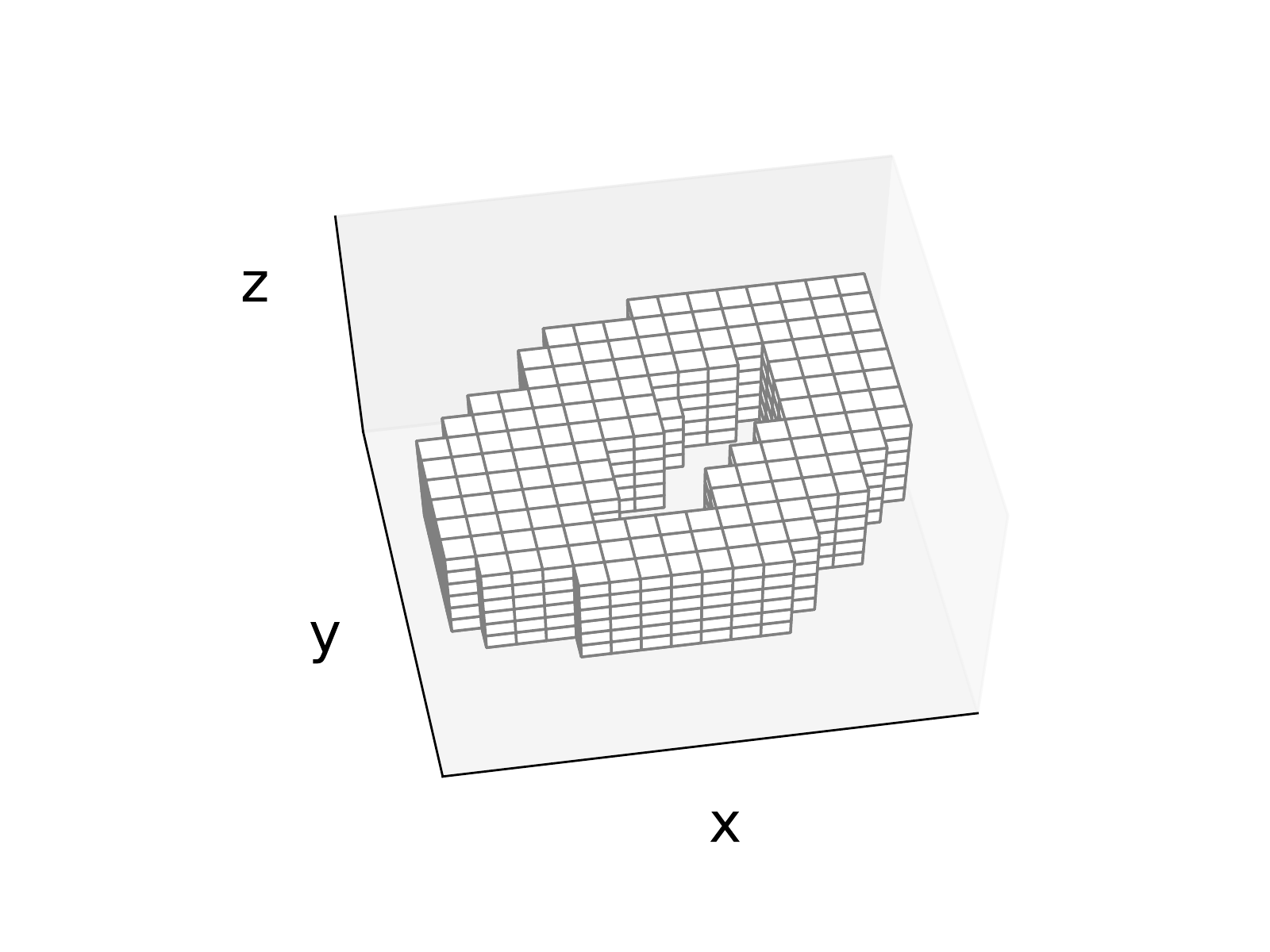} \hfill
    \includegraphics[trim=80 40 80 40, clip,width=0.27\textwidth]{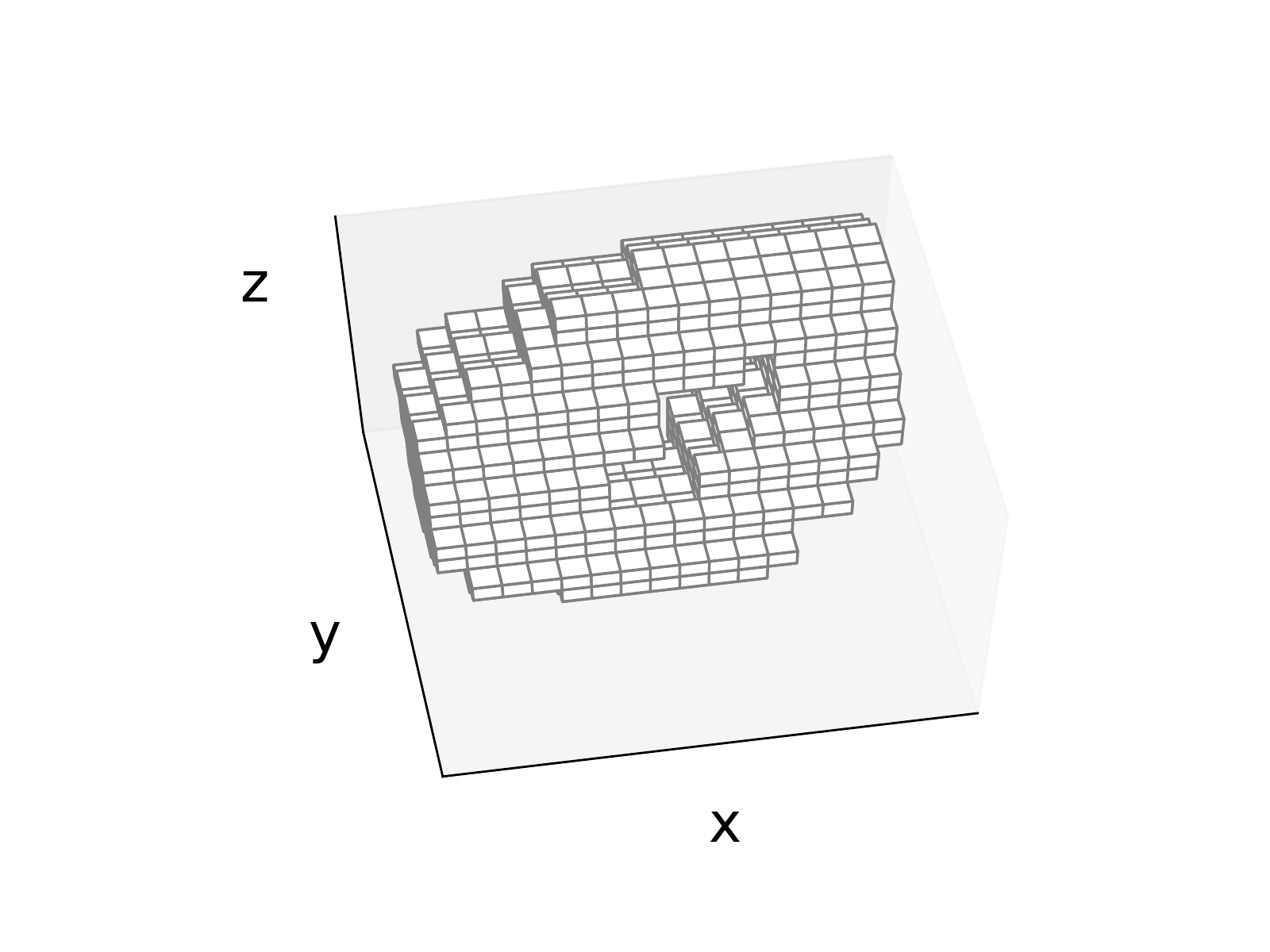} \hfill
    \includegraphics[trim=20 0 40 0, clip,width=0.27\textwidth]{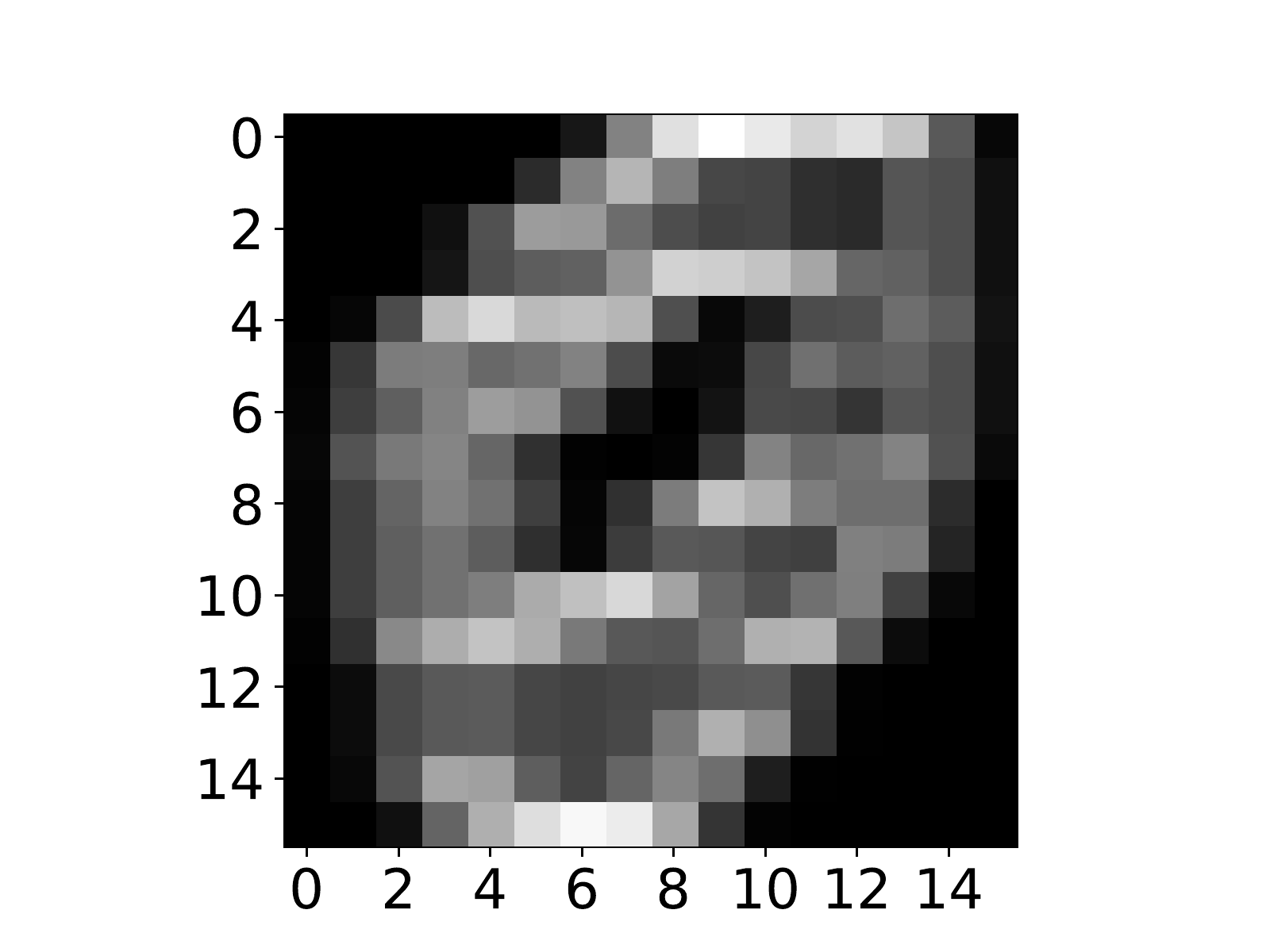}
    \caption{A 3D-MNIST example data point. Left: original input data. Center: input data rotated by $\pi/3$ along the $x$-axis. Right: 2D projection of the rotated data, which is obtained by summation and normalization along the $y$-axis.}
    \label{fig:chap2_3DMNIST_examp}
\end{figure}

\begin{table}[t!]
    \caption{Results for an ensemble of 5 networks trained on 3D-MNIST for independence and a baseline ensemble where each network was trained individually. All results are averaged over 30 runs.}
    \label{tab:chap2_3dmnist}
    \centering
    \scalebox{0.8}{
    \begin{tabular}{c|c|ccccc|c|c}
        \hline
        Loss weight $\lambda$ & & 0 & $10^{-1}$ & $10^{0}$ & $10^{1}$ & $10^{2}$ & Baseline model & Theoretical\\
        \hline\hline
        Mean correlation & & 0.25 & 0.25 & 0.24 & 0.22 & 0.15 & 0.24 & 0.00\\
        \hline\hline
        \multirow{5}{1.8cm}{Single network accuracy} & $k=1$ & 0.70 & 0.70 & 0.70 & 0.66 & 0.38 & 0.70 \\
        & $k=2$ & 0.74 & 0.73 & 0.74 & 0.73 & 0.75 & 0.75 \\
        & $k=3$ & 0.73 & 0.72 & 0.72 & 0.67 & 0.25 & 0.74 \\
        & $k=4$ & 0.81 & 0.80 & 0.79 & 0.73 & 0.17 & 0.82 \\
        & $k=5$ & 0.72 & 0.73 & 0.71 & 0.64 & 0.20 & 0.74 \\
        \hline\hline
        \multirow{5}{1.8cm}{Ensemble accuracy / mean $k$-out-of-5 accuracy} & $k=1$ & 0.70 / 0.97 & 0.70 / 0.97  & 0.70 / 0.97 & 0.66 / 0.96 & 0.38 / 0.88 & 0.70 / 0.97 & 1.00\\
        & $k=2$ & 0.82 / 0.91 & 0.82 / 0.91 & 0.82 / 0.91 & 0.81 / 0.89 & 0.77 / 0.52 & 0.82 / 0.92 & 0.98 \\
        & $k=3$ & 0.84 / 0.81 & 0.84 / 0.80 & 0.84 / 0.80 & 0.83 / 0.75 & 0.78 / 0.22 & 0.85  / 0.82 & 0.90 \\
        & $k=4$ & 0.87 / 0.63 & 0.86 / 0.63 & 0.86 / 0.62 & 0.85/ 0.55 & 0.79 / 0.10 & 0.87 / 0.65 & 0.63 \\
        & $k=5$ & 0.88 / 0.38 & 0.88 / 0.37 & 0.88 / 0.37 & 0.86 / 0.29 & 0.79 / 0.03 & 0.89 / 0.39 & 0.24 \\
        \hline
    \end{tabular}
    }
\end{table}

\begin{figure}[t!]
    \centering
    \includegraphics[trim=20 0 40 0, clip,width=0.49\textwidth]{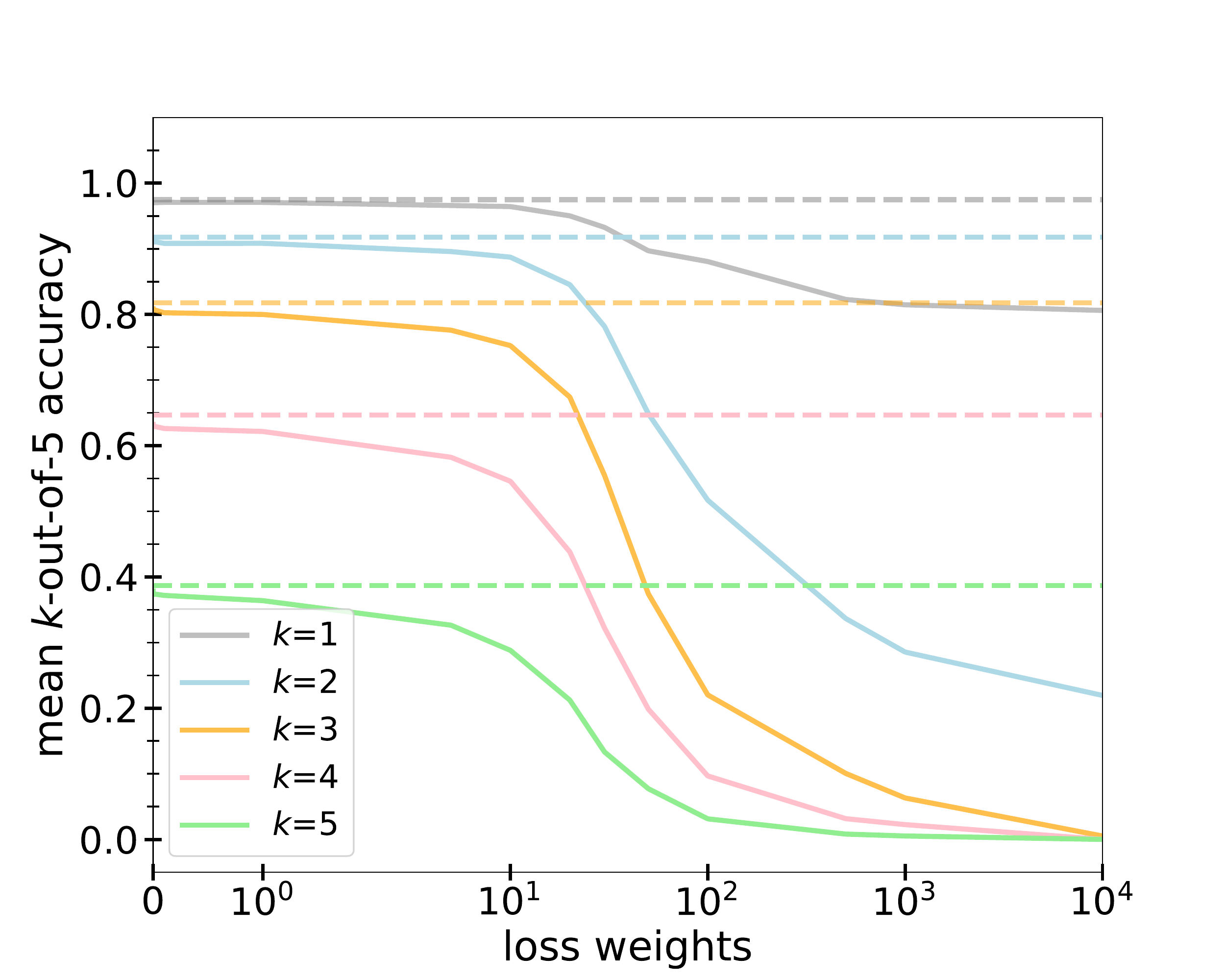}
    \includegraphics[trim=20 0 40 0, clip, width=0.49\textwidth]{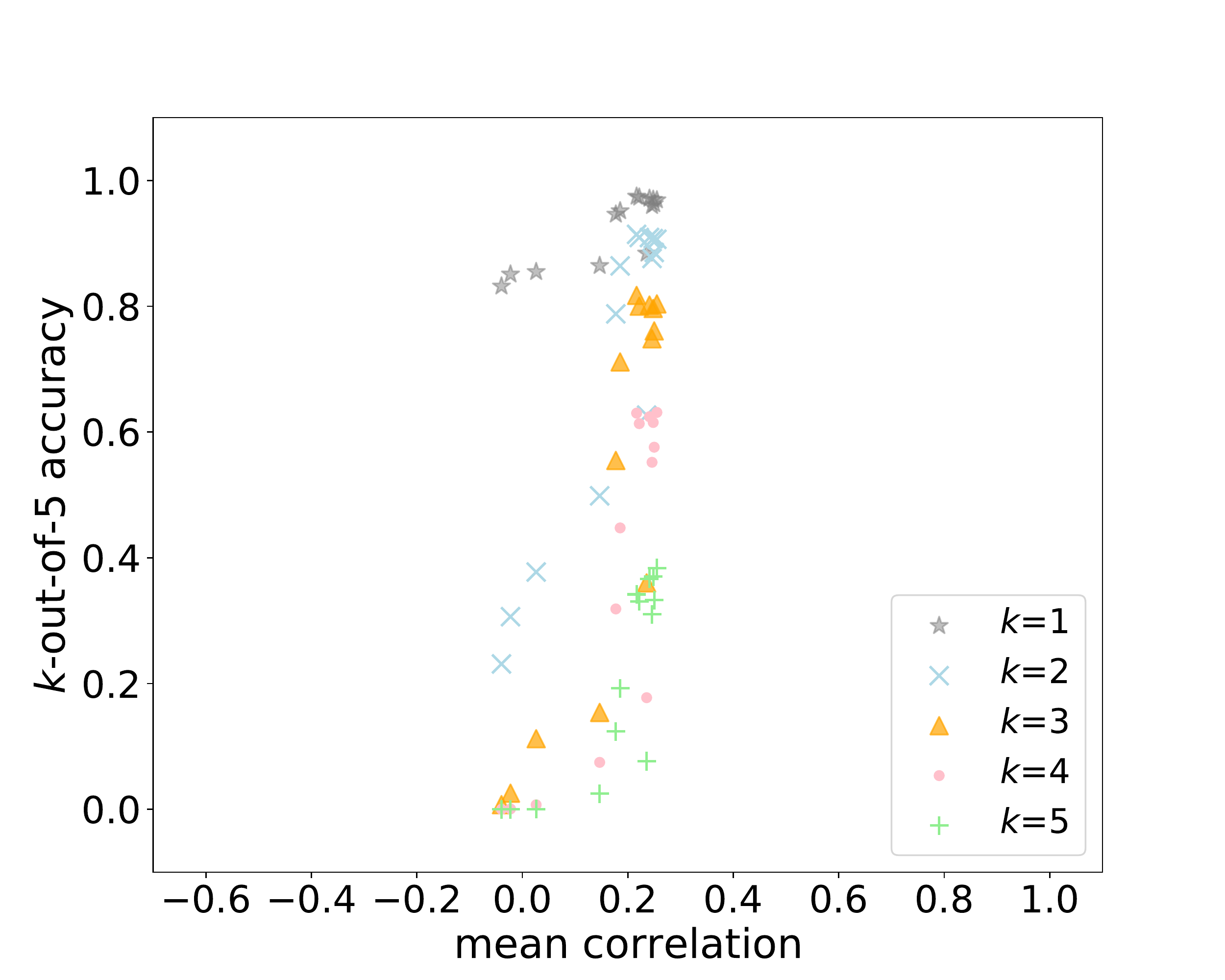}
    \caption{Experiments with the 3D-MNIST dataset. Left: mean $k$-out-of-$5$ accuracy (averaged over 30 repetitions) for an ensemble of five networks as a function of the loss weight $\lambda$. The solid lines depict the accuracies of the ensembles trained for independence with loss weight $\lambda$, the dashed lines depict baseline ensemble accuracies trained without incorporating the loss from~\eqref{eq:chap2_n_networks_inpendence_loss}. Right: $k$-out-of-$5$ accuracy for a single run as a function of the mean correlation (of each network with each other network).}
    \label{fig:chap2_3DMNIST}
\end{figure}

\paragraph{Training of independence for different input sensors:}
In order to conduct further experiments on the scale of the EMNIST and CIFAR10 datasets, we consider the 3D-MNIST dataset that provides synthesized 3D representations of handwritten digits. We apply rotations around the $x$-axis with chosen angles. To this end, we create 5 rotated variants of the dataset with a randomly chosen but fixed angles $\theta = a\pi/9 $, $a=1,\ldots,9$, see Figure \ref{fig:chap2_3DMNIST_examp} for an illustrative example. Each of our $k=1,\ldots,5$ networks obtains one of the 5 rotated variants of the data with a given angle $\theta_k$, and is then trained. For the baseline, all networks are again trained independently. For independence training, all networks are trained with common loss functions, as described previously, and the same handwritten digit, however, viewed from a different angle $\theta_k$, is presented to the network $k$, $k=1,\ldots,5$.

Figure~\ref{fig:chap2_3DMNIST} shows results of numerical experiments conducted analogously to those presented in Figure~\ref{fig:chap2_fivenetworks1}. Comparing both figures with each other, we observe that the baseline $k$-out-of-$5$ accuracies are much lower than when presenting identical images to the networks. Indeed, although MNIST classification constitutes a much simpler task than classifying the letters from EMNIST, this 3D variant shows much lower individual network performances, not only indicated by the left-hand panel of Figure~\ref{fig:chap2_3DMNIST}, but also by Table~\ref{tab:chap2_3dmnist}. On the other hand, the highest mean correlation (obtained for the loss weight $\lambda=0$) for 3D-MNIST depicted by the right-hand panel of Figure~\ref{fig:chap2_3DMNIST} is below $0.3$ and therefore much lower than the one depicted by Figure~\ref{fig:chap2_fivenetworks1}.

Neglecting the reduced performance on 3D-MNIST, these results show that an ensemble, wherein each network obtains a different input, can be clearly improved by increasing the number of ensemble members. The networks only exhibit small correlations among each other. However, it seems that independence training cannot contribute to the performance of the ensemble anymore in the given setup. It remains open, whether independence training may help in a realistic setting with different sensors for perception in automated driving and networks conducting way more difficult tasks such as object detection, instance segmentation, semantic segmentation, or panoptic segmentation. Note that the correlation strengths reported in our experiments in accordance to Section~\ref{sec:Statistical} do not suffice to substantially reduce the data requirements into a feasible regime. Recalling the discussion in Section~\ref{sec:RedndantButCorrelated}, even if the subsystem performance remained unaffected by independence training, based on the lowest correlation observed in our experiments, we could at most expect a reduction of the required data amount by one order of magnitude. 

\section{Conclusion and Outlook}
\label{sec:WayOut}


\paragraph{Summary and main take-away messages:}
In this work, we argued that obtaining statistical evidence from brute-force testing leads to the requirement on infeasible amounts of data. In Sections~\ref{sec:BackOfEnvelope} and~\ref{sec:Statistical} we estimated upper and lower bounds on the amount of data required to test a given perception system with statistical validity for being significantly safer than a human driver. We restricted our considerations to fatalities and arrived at data amounts that are infeasible from both storage and labeling cost perspective, as already found in \cite{kalra2016driving}. In Section~\ref{sec:MeasureLowProbabilities} we showed that perfectly uncorrelated redundant AI perception systems could be used to resolve the problem. 
However, as we have seen in Section~\ref{sec:RedndantButCorrelated}, redundant subsystems that are not perfectly uncorrelated require extremely low correlation coefficients between error events produced by neural networks to yield substantial reductions in data requirement. In Section~\ref{sec:evidence_low_corr} we furthermore found the amount of data needed to prove a sufficiently low correlation as large as the amount of data needed for the original test.

In Section \ref{sec:Redundancy} we present numerical results on the correlation of redundant subsystems. We studied correlation coefficients between error events of redundant neural networks dependent on network architecture, training data and weight initializers. Furthermore, we trained ensembles of neural networks to become decorrelated. Besides studying correlation coefficients, we considered the system's performance in terms of $1$-out-of-$n$ as well as $k$-out-of-$n$ performance. In the numerical experiments we obtained correlation coefficients that would allow for a reduction in the amount of data required by at most one order of magnitude.  For the testing problem, this would  still represent an infeasible data requirement.
However, redundancy could contribute to a moderate reduction of the amount of data required for testing and potentially be combined with other approaches.  

There are alternative approaches to test the safety of perception systems other than brute-force testing. Here, we give a short outlook on two of them that are very actively developed in the research community.

\paragraph{Outlook on testing with synthetic data:}
One possibility to obtain vast amounts of data for testing is to consider synthetic sources of data. The question, whether synthetic data can be used for testing has already been addressed, e.g., in \cite{rosenzweig2021validation}. In principle, arbitrary amounts of test data can be created from a driving simulation such as CARLA \cite{dosovitskiy2017carla}. The domain shift between synthetic and real data can be bridged with generative adversarial networks (GANs). The latter have been proven to be learnable in the large sample limit \cite{biauGAN,asatryan2020gan}, meaning that for increasing amounts of data and capacity of the generator and discriminator, the learned map from synthetic to real data is supposed to converge to the true one. Combining synthetic data and adversarial learning is therefore a promising candidate for testing DNNs. However, in this setup there remain other gaps to be bridged (number of assets, environmental variability, and infeasibility of the empirically risk minimizing generator). 

\paragraph{Outlook on testing with real data:}
There exists a number of helpful approaches to estimate the performance on unlabeled data.
A DNN of strong performance (or ensembles of those) can be utilized to compute pseudo ground truth. The model to be equipped in an AI perception system can be learned in a teacher-student fashion \cite{teacherstudent2019,Baer2019}, and the discrepancies between the student model and the teachers can be compared with errors on a moderate subset with ground truth, for instance in terms of correlations of errors and discrepancies.
Furthermore, in order to process the vast amounts of recorded data and perform testing more efficiently, well performing uncertainty quantification methods \cite{kendall2017uncertainties,kohl2018probabilistic,Rottmann18} in combination with corner case detection methods \cite{bolte:2019,heidecker2021application} can help to pre-select data for testing. Besides that, many additional approaches towards improving the reliability of DNN exist. However, while a big number of tools already exist, their proper application to DNN testing and inference of statistically relevant statements on the system's safety still requires thorough research.

These approaches and other upcoming research might be part of the solution of the testing problem in future.  

\textbf{Concluding remark}:
As a concluding remark, this article does not intend to discourage safety arguments, as also conceptualized in this volume.  We do not deny the value of empirical evidence in order to, \eg prove the relative superiority of one AI system over another with regards to safety.  The inherent difficulty to provide direct evidence for the better-than-human safety of automated driving as required by the ethics committee of the German Ministry of Transportation and Digital Infrastructure \cite{fabio2017ethik} should not be mistaken as an excuse for a purely experimental approach. Bringing automated vehicles to the street without prior risk assessment implies that risks would be judged \emph{a posteriori} based on the experience with a large fleet of automated vehicles. 

Such matters attain urgency in the light of recent German legislation on the experimental usage of automated driving under human supervision\footnote{\url{https://dserver.bundestag.de/btd/19/274/1927439.pdf}} which only refers to the technical equipment for the sensing of automated vehicles, but does not specify the minimal performance for the AI-based perception based on the sensor information.   Related regulations in other countries face similar problems\footnote{ Framework for Automated Driving System Safety, No. NHTSA-2020-0106, 49 CFR Part 571 (Nov. 19, 2020).}. 

The debate, how to ensure a safe transition to automated driving that complies with high ethical standards,  therefore remains of imminent scientific and public interest.          

\subsection*{Acknowledgments} The authors thank Lina Haidar for support and numerical results in Section 4.2. Financial support by the German Federal Ministry of Economic Affairs and Energy (BMWi) via Grant No. 19A19005R as a part of the Safe AI for Automated Driving consortium is gratefully acknowledged.


\setnewpage
\putbib[references] 
\end{bibunit}\setnewpage
\fi


\ifchapthree
\begin{bibunit}[\bibstylenew]
\setcounter{section}{1}
\setcounter{figure}{0}
\setcounter{table}{0}
\setcounter{equation}{0}
\setcounter{algorithm}{0}
\addchap{Analysis and Comparison of Datasets by Leveraging Data Distributions in Latent Spaces}\label{sec:analysis_and_comparison}
\addtocontents{toc}{Hanno Stage, Lennart Ries, Jacob Langner, Stefan Otten, and Eric Sax \protect\vspace{0.2em}\par}
\chapterauthor{
Hanno Stage\footnotemark[1],
Lennart Ries\footnotemark[1],
Jacob Langner\footnotemark[1],
Stefan Otten\footnotemark[1],
and Eric Sax\footnotemark[1]
}
\footnotetext[1]{FZI Research Center for Information Technology, Haid-und-Neu-Str.~10-14, 76131 Karlsruhe, Germany, \{stage, ries, langner, otten, sax\}@fzi.de}
\setcounter{footnote}{1}
\input{part2/chapter_3/stage_chap3}\setnewpage
\putbib[references]
\end{bibunit}\setnewpage
\fi


\ifchapfour
\begin{bibunit}[\bibstylenew]
\setcounter{section}{1}
\setcounter{figure}{0}
\setcounter{table}{0}
\setcounter{equation}{0}
\setcounter{algorithm}{0}
\addchap{Optimized Data Synthesis for DNN Training and Validation by Sensor Artifact Simulation}\label{sec:improving_the_performance}
\addtocontents{toc}{Korbinian Hagn and Oliver Grau \protect\vspace{0.2em}\par}
\chapterauthor{
Korbinian Hagn\footnotemark[1]
and Oliver Grau\footnotemark[1]
}
\footnotetext[1]{Intel Deutschland GmbH, Lilienthalstraße 15, 85579 Neubiberg, Germany, \{korbinian.hagn, oliver.grau\}@intel.com}
\input{part2/chapter_4/hagn_chap4}\setnewpage
\putbib[references] 
\end{bibunit}\setnewpage
\fi


\ifchapfive
\begin{bibunit}[\bibstylenew]
\setcounter{section}{1}
\setcounter{figure}{0}
\setcounter{table}{0}
\setcounter{equation}{0}
\setcounter{algorithm}{0}
\addchap{Improved DNN Robustness by Multi-Task Training With an Auxiliary Self-Supervised Task}\label{sec:improved_dnn_robustness}
\addtocontents{toc}{Marvin Klingner and Tim Fingscheidt \protect\vspace{0.2em}\par}
\chapterauthor{
Marvin Klingner\footnotemark[1]
and Tim Fingscheidt\footnotemark[1]
}
\footnotetext[1]{Technische Universität Braunschweig, Institute for Communications Technology (IfN), Schleinitzstr.\ 22, 38106 Braunschweig, Germany, \{m.klingner, t.fingscheidt\}@tu-bs.de}
\setcounter{footnote}{1}
\input{part2/chapter_5/klingner_chap5}\setnewpage
\putbib[references]
\end{bibunit}\setnewpage
\fi


\ifchapsix
\begin{bibunit}[\bibstylenew]
\setcounter{section}{1}
\setcounter{figure}{0}
\setcounter{table}{0}
\setcounter{equation}{0}
\setcounter{algorithm}{0}
\addchap{Improving Transferability of Generated Universal Adversarial Perturbations for Image Classification and Segmentation}\label{sec:uap}
\addtocontents{toc}{Atiye Sadat Hashemi, Andreas Bär, Saeed Mozaffari, and Tim Fingscheidt \protect\vspace{0.2em}\par}
\chapterauthor{
Atiye Sadat Hashemi$^{\textrm{1,2}}$,
Andreas Bär$^{\textrm{1}}$,
Saeed Mozaffari$^{\textrm{2,3}}$,
and Tim Fingscheidt$^{\textrm{1}}$
}
\footnotetext[1]{Technische Universität Braunschweig, Institute for Communications Technology (IfN), Schleinitzstr.~22, 38106 Braunschweig, Germany, \{andreas.baer, t.fingscheidt\}@tu-bs.de, atiye.hashemi1991@gmail.com}
\footnotetext[2]{Semnan University, Campus 1, 35131-19111, Semnan, Iran, \{atiye.hashemi, mozaffari\}@semnan.ac.ir}
\footnotetext[3]{University of Windsor, 401 Sunset Ave, Windsor, ON N9B 3P4, Canada, saeed\_mozaffari@yahoo.com}
\setcounter{footnote}{3}
\input{part2/chapter_6/hashemi_chap6}\setnewpage
\putbib[references] 
\end{bibunit}\setnewpage
\fi


\ifchapseven
\begin{bibunit}[\bibstylenew]
\setcounter{section}{1}
\setcounter{figure}{0}
\setcounter{table}{0}
\setcounter{equation}{0}
\setcounter{algorithm}{0}
\addchap{Invertible Neural Networks for Understanding Semantics of Invariances of CNN Representations}\label{sec:invertible_neural_networks}
\addtocontents{toc}{Robin Rombach, Patrick Esser, Andreas Blattmann, and Björn Ommer \protect\vspace{0.2em}\par}
\chapterauthor{
Robin Rombach\footnotemark[1],
Patrick Esser\footnotemark[1],
Andreas Blattmann\footnotemark[1],
and Björn Ommer\footnotemark[1]
}
\footnotetext[1]{HCI, IWR, Heidelberg University, Berliner Str. 43, 69120 Heidelberg, Germany, \{robin.rombach, patrick.esser, andreas.blattmann, bjoern.ommer\}@iwr.uni-heidelberg.de}
\setcounter{footnote}{1}
\input{part2/chapter_7/rombach_chap7-commands}
\input{part2/chapter_7/rombach_chap7-tables}
\input{part2/chapter_7/rombach_chap7-figures}
\input{part2/chapter_7/rombach_chap7}\setnewpage
\putbib[references]
\end{bibunit}\setnewpage
\fi


\ifchapeight
\begin{bibunit}[\bibstylenew]
\setcounter{section}{1}
\setcounter{figure}{0}
\setcounter{table}{0}
\setcounter{equation}{0}
\setcounter{algorithm}{0}
\addchap{Confidence Calibration for Object Detection and Segmentation}\label{sec:confidence_calibration_for}
\addtocontents{toc}{Fabian Küppers, Anselm Haselhoff, Jan Kronenberger, 
and Jonas Schneider \protect\vspace{0.2em}\par}
\chapterauthor{
Fabian Küppers\footnotemark[1],
Anselm Haselhoff\footnotemark[1],
Jan Kronenberger\footnotemark[1],
and Jonas Schneider\footnotemark[2]
}
\footnotetext[1]{Hochschule Ruhr West, Duisburger Str.~100, 45479 Mülheim a.d.~Ruhr, Germany, \{fabian.kueppers, anselm.haselhoff, jan.kronenberger\}@hs-ruhrwest.de}
\footnotetext[2]{Elektronische Fahrwerksysteme GmbH, Dr.-Ludwig-Kraus-Str.~6, 85080 Gaimersheim, Germany, jonas.schneider@efs-auto.com}
\setcounter{footnote}{2}
\input{part2/chapter_8/kueppers_chap8}\setnewpage
\putbib[references]
\end{bibunit}\setnewpage
\fi


\ifchapnine
\begin{bibunit}[\bibstylenew]
\setcounter{section}{1}
\setcounter{figure}{0}
\setcounter{table}{0}
\setcounter{equation}{0}
\setcounter{algorithm}{0}
\addchap{Uncertainty Quantification for Object Detection: Output- and Gradient-based Approaches}\label{sec:uncertainty_qunatification_for}
\addtocontents{toc}{Tobias Riedlinger, Marius Schubert, Karsten Kahl, and Matthias Rottmann \protect\vspace{0.2em}\par}
\chapterauthor{
Tobias Riedlinger\footnotemark[1],
Marius Schubert\footnotemark[1],
Karsten Kahl\footnotemark[1],
and Matthias Rottmann\footnotemark[1]
}
\footnotetext[1]{University of Wuppertal, Gaußstr. 20, 42119 Wuppertal, Germany, \{riedlinger, schubert, kkahl, rottmann\}@uni-wuppertal.de}
\setcounter{footnote}{1}
\input{part2/chapter_9/riedlinger_chap9}\setnewpage
\putbib[references]
\end{bibunit}\setnewpage
\fi


\ifchapten
\begin{bibunit}[\bibstylenew]
\setcounter{section}{1}
\setcounter{figure}{0}
\setcounter{table}{0}
\setcounter{equation}{0}
\setcounter{algorithm}{0}
\addchap{Detecting and Learning the Unknown in Semantic Segmentation}\label{sec:finding_unknown_objects}
\addtocontents{toc}{Robin Chan, Svenja Uhlemeyer, Matthias Rottmann, and Hanno Gottschalk \protect\vspace{0.2em}\par}
\chapterauthor{
Robin Chan\footnotemark[1],
Svenja Uhlemeyer\footnotemark[1],
Matthias Rottmann\footnotemark[1],
and Hanno Gottschalk\footnotemark[1]
}
\footnotetext[1]{University of Wuppertal, Gaußstr.~20, 42119 Wuppertal, Germany, \{rchan, suhlemeyer, rottmann, hanno.gottschalk\}@uni-wuppertal.de}
\setcounter{footnote}{1}
\input{part2/chapter_10/chan_chap10}\setnewpage
\putbib[references]
\end{bibunit}\setnewpage
\fi


\ifchapeleven
\begin{bibunit}[\bibstylenew]
\setcounter{section}{1}
\setcounter{figure}{0}
\setcounter{table}{0}
\setcounter{equation}{0}
\setcounter{algorithm}{0}
\addchap{Evaluating Mixture-of-Expert Architectures for Network Aggregation}\label{sec:evaluating_mixture_of}
\addtocontents{toc}{Svetlana Pavlitskaya, Christian Hubschneider, and Michael Weber \protect\vspace{0.2em}\par}
\chapterauthor{
Svetlana Pavlitskaya\footnotemark[1],
Christian Hubschneider\footnotemark[1],
and Michael Weber\footnotemark[1]
}
\footnotetext[1]{FZI Research Center for Information Technology, Haid-und-Neu-Str.~10-14, 76131 Karlsruhe, Germany, \{pavlitskaya, hubschneider, weber\}@fzi.de}
\setcounter{footnote}{1}
\input{part2/chapter_11/pavlitskaya_chap11}\setnewpage
\putbib[references] 
\end{bibunit}\setnewpage
\fi

 

\ifchaptwelve
\begin{bibunit}[\bibstylenew]
\setcounter{section}{1}
\setcounter{figure}{0}
\setcounter{table}{0}
\setcounter{equation}{0}
\setcounter{algorithm}{0}
\addchap{Safety Assurance of Machine Learning for Perception Functions}\label{sec:evidences_for_the}
\addtocontents{toc}{Simon Burton, Christian Hellert, Fabian Hüger, Michael Mock, and Andreas Rohatschek \protect\vspace{0.2em}\par}
\chapterauthor{
Simon Burton\footnotemark[1],
Christian Hellert\footnotemark[2],
Fabian Hüger\footnotemark[3],
Michael Mock\footnotemark[4],
and Andreas Rohatschek\footnotemark[5]
}
\footnotetext[1]{Fraunhofer Institute for Cognitive Systems IKS, Hansastraße 32, 80686 Munich, Germany, simon.burton@iks.fraunhofer.de}
\footnotetext[2]{Continental AG, Bessie-Coleman-Str. 7, 60549 Frankfurt am Main, Germany, christian.hellert@continental.com}
\footnotetext[3]{Volkswagen AG, Berliner~Ring~2, 38440~Wolfsburg, Germany, fabian.hueger@volkswagen.de}
\footnotetext[4]{Fraunhofer Institute for Intelligent Analysis and Information Systems IAIS, Schloss Birlinghoven 1, 53757 Sankt Augustin, Germany, michael.mock@iais.fraunhofer.de}
\footnotetext[5]{Robert Bosch GmbH, Robert-Bosch-Campus 1, 71272 Renningen, Germany, andreas-juergen.rohatschek@de.bosch.com}
\setcounter{footnote}{5}
\input{part2/chapter_12/burton_chap12}\setnewpage
\putbib[references]
\end{bibunit}\setnewpage
\fi


\ifchapthirteen
\begin{bibunit}[\bibstylenew]
\setcounter{section}{1}
\setcounter{figure}{0}
\setcounter{table}{0}
\setcounter{equation}{0}
\setcounter{algorithm}{0}
\addchap{A Variational Deep Synthesis Approach for Perception Validation}\label{sec:highly_variational_validation}
\addtocontents{toc}{Oliver Grau, Korbinian Hagn, and Qutub Syed Sha \protect\vspace{0.2em}\par}
\chapterauthor{
Oliver Grau\footnotemark[1], Korbinian Hagn\footnotemark[1],
and Qutub Syed Sha\footnotemark[1]
}
\footnotetext[1]{Intel Deutschland GmbH, Lilienthalstraße 15, 85579 Neubiberg, Germany, \{oliver.grau, korbinian.hagn, syed.qutub\}@intel.com}
\setcounter{footnote}{1}
\input{part2/chapter_13/grau_chap13}\setnewpage
\putbib[references]
\end{bibunit}\setnewpage
\fi


\ifchapfourteen
\begin{bibunit}[\bibstylenew]
\setcounter{section}{1}
\setcounter{figure}{0}
\setcounter{table}{0}
\setcounter{equation}{0}
\setcounter{algorithm}{0}
\addchap{The Good and the Bad: Using Neuron Coverage as a DNN Validation Technique}\label{sec:the_good_and}
\addtocontents{toc}{Sujan Sai Gannamaneni, Maram Akila, Christian Heinzemann, and Matthias Woehrle \protect\vspace{0.2em}\par}
\chapterauthor{
Sujan Sai Gannamaneni\footnotemark[1],
Maram Akila\footnotemark[1],
Christian Heinzemann\footnotemark[2],
and Matthias Woehrle\footnotemark[2]
}
\footnotetext[1]{Fraunhofer Institute for Intelligent Analysis and Information Systems IAIS, Schloss Birlinghoven 1, 53757 Sankt Augustin, Germany, \{sujan.sai.gannamaneni, maram.akila\}@iais.fraunhofer.de}
\footnotetext[2]{Robert Bosch GmbH, Corporate Research, Robert-Bosch-Campus 1, 71272 Renningen, Germany, \{christian.heinzemann, matthias.woehrle\}@de.bosch.com}
\setcounter{footnote}{2}
\input{part2/chapter_14/gannamaneni_chap14}\setnewpage
\putbib[references]
\end{bibunit}\setnewpage
\fi


\ifchapfifteen
\begin{bibunit}[\bibstylenew]
\setcounter{section}{1}
\setcounter{figure}{0}
\setcounter{table}{0}
\setcounter{equation}{0}
\setcounter{algorithm}{0}
\addchap{Joint Optimization for DNN Model Compression and Corruption Robustness}\label{sec:achieving_model_compression}
\addtocontents{toc}{Serin Varghese, Christoph Hümmer, Andreas Bär, Fabian Hüger, and Tim Fingscheidt \protect\vspace{0.2em}\par}
\chapterauthor{
Serin Varghese\footnotemark[1],
Christoph Hümmer\footnotemark[1],
Andreas Bär\footnotemark[2],
Fabian Hüger\footnotemark[1],
and Tim Fingscheidt\footnotemark[2]
}
\footnotetext[1]{Volkswagen AG, Berliner~Ring~2, 38440~Wolfsburg, Germany, \{john.serin.varghese, christoph.heummer, fabian.hueger\}@volkswagen.de}
\footnotetext[2]{Technische Universität Braunschweig, Institute for Communications Technology (IfN), Schleinitzstr.~22, 38106 Braunschweig, Germany, \{andreas.baer, t.fingscheidt\}@tu-bs.de}
\setcounter{footnote}{2}
\input{part2/chapter_15/varghese_chap15}\setnewpage
\putbib[references] 
\end{bibunit}\setnewpage
\fi


\end{document}
